\documentclass[
]{ceurart}

\usepackage{listings}
\usepackage{float}
\usepackage{newfloat}
\usepackage{caption}
\DeclareCaptionStyle{ruled}{labelfont=normalfont,labelsep=colon,strut=off} % DO NOT CHANGE THIS
\floatstyle{ruled}
\newfloat{listing}{tb}{lst}{}
\floatname{listing}{Listing}
\lstdefinelanguage{json}{}

\usepackage[most]{tcolorbox} % for colored boxes

\newcommand{\R}{\mathbb{R}}

\newcommand{\calG}{\mathcal{G}}
\newcommand{\calH}{\mathcal{H}}

\newcommand{\calU}{\mathcal{U}}

\newcommand{\calI}{\mathcal{I}}

\newcommand{\hatr}{\hat{r}}

\newcommand{\EVOI}{\mathrm{EVOI}}
\newcommand{\PEU}{\mathrm{PEU}}
\newcommand{\EU}{\mathrm{EU}}

\newtcolorbox{mmbox}[1]{
  colback=white,
  colframe=blue!75!black,
  fonttitle=\bfseries,
  fontupper=\small,
  title=#1,
  boxsep=0cm,
}

\newcommand{\icer}{\emph{ICER}}

\begin{document}
\copyrightyear{2025}
\copyrightclause{Copyright for this paper by its authors.
  Use permitted under Creative Commons License Attribution 4.0
  International (CC BY 4.0).}
\conference{GENNEXT@SIGIR'25: The 1st Workshop on Next Generation of IR and Recommender Systems with Language Agents, Generative Models, and Conversational AI, Jul 17, 2025, Padova, Italy}
\title{Synthetic Dialogue Generation for Interactive Conversational Elicitation \& Recommendation (ICER)}

\author[1]{Moonkyung Ryu}[%
orcid=0009-0000-1088-9139,
email=mkryu@google.com,
]
\cormark[1]
\fnmark[1]
\address[1]{Google Research, Mountain View, California, USA}

\author[1]{Chih-Wei Hsu}[%
orcid=0000-0001-9929-9951,
email=cwhsu@google.com,
]
\fnmark[1]

\author[1]{Yinlam Chow}[%
orcid=0000-0002-7533-8300,
email=yinlamchow@google.com,
]

\author[2]{Mohammad Ghavamzadeh}[%
orcid=0000-0003-0930-8688,
email=ghavamza@amazon.com,
]
\address[2]{Amazon AGI, California, USA}

\author[1]{Craig Boutilier}[%
orcid=0000-0001-9330-4545,
email=cboutilier@google.com,
]

\cortext[1]{Corresponding author.}
\fntext[1]{These authors contributed equally.}

\begin{abstract}
While language models (LMs) offer great potential for conversational recommender systems (CRSs), the paucity of public CRS data makes fine-tuning LMs for CRSs challenging. In response, LMs as \emph{user simulators qua data generators} can be used to train LM-based CRSs,
but often lack \emph{behavioral consistency}, generating utterance sequences inconsistent with those of \emph{any} real user. To address this, we develop a methodology for generating natural dialogues that are consistent
with a user's underlying state
using behavior simulators together with \emph{LM-prompting}.
We illustrate our approach by generating a large, open-source CRS data set with both preference elicitation and example critiquing. Rater evaluation on 
some of these dialogues shows them to exhibit considerable consistency, factuality and naturalness. 
\end{abstract}


\maketitle

\section{Introduction}
Among \emph{recommender systems (RSs)},
% are deeply ingrained in many facets of daily life, and 
\emph{conversational RSs (CRSs)} have emerged as a means to provide more natural, higher-bandwidth communication with users \cite{radlinski_EtAl:aaai22}.
% The development of 
Sophisticated \emph{language models (LMs)}, e.g., Gemini \citep{team2023gemini}, GPT \citep{achiam2023gpt}, offer even greater potential for a holistic approach to CRSs that supports preference elicitation, user critiquing, exploration, explanation
%, means-ends analysis 
and more. Unfortunately, the paucity of open-ended public conversational data means that there is little data on which to train/fine-tune 
% the required
personalized LMs.

As a result, 
% a number of researchers 
some
have proposed using LMs as \emph{user simulators} to interact with an LM-based CRS to generate useful conversational 
% data for training or fine-tuning the LM driving a CRS.
training data.
\emph{``LMs qua simulators qua data generators''} of this form are often prompted to behave as if it were a specific type of user by the inclusion of a \emph{persona}, \emph{preference profile} or other such signals in the prompt itself \cite{park23}.\footnote{Indeed, the use of LMs as user simulators in a variety of tasks has become commonplace \cite{park23,gui23,shapira24}.} Alternatives include \emph{ELM} \cite{tennenholtz2024demystifying}, which uses LMs to interpret user embeddings, and \emph{UserSimCRS}, \cite{user_sim_crs} which employs a knowledge graph for user preferences.

While promising, LM-based simulators suffer from a host of problems. One of the more pressing is \emph{behavioral consistency}: (1) Do individual behavior trajectories generated by the prompted LM correspond to plausible behaviors of human users in the domain of interest? (2) Does the distribution of such behaviors reflect 
% the distribution of real user behaviors in 
that of
the population of interest? We might target a variety of behavioral characteristics in an LM-based simulator, but here we focus on one in particular, \emph{preference consistency}: are the behaviors (and corresponding utterances) of a simulated user consistent with a specific preference (or utility) function over the recommendation domain, and is the behavioral distribution reflective of the real-world preference distribution of interest?

In this work, we propose a methodology to ensure preference consistency in LM-based user simulation, and illustrate 
it
% its realization
with a particular CRS. We assume access to a user simulator grounded in suitable behavioral models and data. In our case, multi-turn user interactions with a (non-conversational) RS are generated using existing, well-justified user choice and response models, with user preferences drawn from a distribution over user preference embeddings. Generated user interactions include recommendation acceptance/rejection, example critiques, and response to RS preference elicitation queries---each interaction in a trajectory is (noisily) generated w.r.t.\ the sampled user embedding. To generate conversational data, instead of simple templates to translate (user, RS) actions into a natural language (NL) utterance---which leads to unnatural, repetitive dialogues---we use \emph{LM-prompting}, employing an LM to further refine the templatized responses and inject natural variation.

The three stages---behavior generation, templatized NL construction, LM-based utterance refinement---comprise a user simulator of \emph{Interactive Conversational Elicitation for Recommendation (\icer)} that generates behaviorally grounded dialogues exhibiting individual preference consistency. Using the \icer\ simulator, we generate \emph{MD-DICER}, a dataset of 100K CRS dialogues based on MovieLens 25M \cite{harper16:movielens}, and critical metadata (e.g., user embeddings for each dialogue, item embeddings, etc.).
% \footnote{MD-DICER is openly available at \textbf{https://github.com/google-research-datasets/md-dicer}}
\footnote{MD-DICER: https://www.kaggle.com/datasets/googleai/md-dicer}
Using rater evaluation, we assess the plausibility and quality of the generated dialogues, showing them to be more natural, fluent, factual and consistent than their templatized counterparts. We further evaluate these dialogues, both complete and partial prefixes, serving as prompts to off-the-shelf LMs, and demonstrate that they provide the LM with a deeper understanding of a user than the corresponding ``prior.'' While we illustrate the \icer\ methodology using a specific behavioral RS and data set, we believe that it (and the MD-DICER conversations) should spur further research into LM-based CRSs and behaviorally consistent LM-based simulation.

\section{Preliminaries}
\label{sec:prelim}

We first outline our problem formulation and assumptions and briefly discuss related work.

% \paragraph{Recommender Systems}

\vspace*{1mm}
\noindent
\textbf{Recommender Systems.}
To model user preferences over items in an RS, we assume a standard collaborative filtering (CF) model \citep{su2009survey}, where user behavioral data (e.g., clicks, ratings, etc.) is used to learn user and item \emph{representations}.
% which are combined to predict a user's preferences for a given item.
We assume users $\mathcal U$, items $\mathcal I$, and a (usually sparse) $|\mathcal I|\times |\mathcal U|$ ratings matrix~$\mathcal R = \{(u,i,r_{u,i}) : r_{u,i} \neq 0\}$,
where $r_{u,i}$ is user~$u$'s rating of item $i$. A CF method learns both user and item \emph{embeddings} from $\mathcal R$, where the embedding representation
$\phi_I: \calI \rightarrow \R^d$ of $i\in{\mathcal I}$ captures its latent attributes, and
the representation $\phi_U:\calU \rightarrow \R^d$ of $u\in{\mathcal U}$ reflects their \emph{utility function} over these attributes, and the user-item affinity is estimated via $\hatr_{i,u} = \phi_U(u)^\top \phi_I(i)$. We use this user embedding to drive user simulated responses, but assume it is unknown to the RS.
% To capture the preference variations of users reflected from their noisy responses,
The RS's uncertainty over  $\phi_u$ is captured by a \emph{prior belief}, or distribution $P_U:\calU\rightarrow \mathbb{P}(\mathbb R^d)$, learned via probabilistic matrix factorization~\cite{salakhutdinov-mnih:nips07}.

\vspace*{1mm}
\noindent
\textbf{Soft Attributes.}
Effective CRSs should allow natural, multi-turn NL-based interaction with a user. Specifically, the user should be able to use NL to communicate their preferences over item attributes, critique recommended items, and answer CRS-initiated preference elicitation queries. Moreover, the open-ended nature of CRSs should support the use of \emph{soft attributes} \citep{radlinski:sdd2019}, item attributes that are not part of an agreed-upon, formal item specification, since these are often the way in which users conceive of their preferences. For example, users might describe movies (and their preferences) using terms like `funny,' `thought-provoking,' `violent,' `cheesy,' etc.

While CF methods have been widely used to predict user-item affinity based on past behavior, their item representations are complex and opaque, which makes
NL-based interaction challenging. 
Research on interpretable representations has been used to render non-transparent ML-based representations more understandable \citep{sundararajan:icml2017}.
Recently, \emph{concept activation vectors (CAVs)}
\citep{kimTCAV:icml18} have been applied by \citet{gopfert2021discovering} in RSs to identify the semantics of soft attributes---in this case, informal tags applied by users---w.r.t.\ the item representation learned by CF models. Using this approach, a CAV $c_g$ is a direction in embedding space that predicts the degree $c_g^\top\phi_I(i)$ to which an item $i$ exhibits an attribute $g$. These CAVs are learned using logistic regression or pairwise ranking over informal, often subjective, tag/attribute usage \emph{on top of the learned item representations} $\phi_I(i)$.

% is computed to determine if the CF model has learned a representation of each tag $g$. Let $T_g = \{(u,i) : t_{u,i,g}\!=\! 1\}$ be the set of user-item pairs for which tag $g$ has been applied. % One way to learn this CAV is via (regularized) logistic regression:
% $\min_{c_g}\sum_{(\cdot,i)\in T_g} \log(1 + e^{-c_g^\top \phi_I(i)}) + \frac{\lambda}{2} c_g^\top c_g$.
% %
% Notice that CAV $c_g$ is the normal to the separating hyperplane of this classifier, offering a directional semantics for tag $g$ in the item embedding space, i.e., $c_g^\top\phi_I(i)$ quantifies the degree to which item $i$ satisfies tag $g$. 

% \paragraph{Dialogue Inpainting}

\vspace*{1mm}
\noindent
\textbf{Dialogue Inpainting.}
Dialogue inpainting has emerged as an effective way to generate synthetic conversational data in domains where such organic data is lacking \citep{pmlr-v162-dai22a}. As such it can be incredibly useful for bootstrapping the training of dialogue systems with novel functionality. \citet{pmlr-v162-dai22a} propose inpainting from document data to assist with conversational question answering (QA). Data is generated by treating a document's sentences as dialogue utterances by the ``writer,'' and uses an LM to generate what a hypothetical ``reader'' might have said/asked in between these utterances. 
\citet{leszczynski2023talk} use inpainting to generate CRS data, much like we do here, though they focus on the use of pre-existing curated item sequences (specifically, playlists), treating them much like documents, and using an LM to ``inpaint'' utterances that a user might offer to induce the next item (or small subset) in the sequences. While similar in spirit to our work, this approach does not incorporate system initiative like recommendation or preference elicitation (PE), nor does it systematic inject user responses to PE queries based on a user's underlying preferences.

% can effectively generate large amounts of multi-turn user-agent conversation data. It can take any arbitrary document as input and transforms it into a dialogue between the writer and an imagined reader by treating sentences from the document as utterances spoken by the writer. Then, an LM predicts what the imagined reader might have said in between each of the writer's utterances. Leveraging the LM's ability to understand and respond to the context, dialogue inpainting offers the potential to create more nuanced and engaging conversations compared to simpler template-based approaches. 

% \paragraph{Other Related Work}

\vspace*{1mm}
\noindent
\textbf{Other Related Work.}
Synthetic data provides a compelling solution to the challenges of privacy preservation and data scarcity often faced in the development of conversational recommender (ConvRec) systems.  Traditional ConvRec systems rely heavily on historical user-agent dialogue data, which not only tends to be limited in quantity and quality but also carries the inherent risk of compromising sensitive user information.
Previous synthetic data generation methods rely on explicit probabilistic models, either manually configured or designed to mirror aggregate statistics from historical sources \cite{DELCARMENRODRIGUEZHERNANDEZ2017516}. However, these approaches primarily generate attributes and ratings, not the NL utterances of focus in our work. Some studies generate synthetic CRS data using item ratings \cite{DBLP:journals/corr/DodgeGZBCMSW15}, text reviews \cite{10.1145/3269206.3271776}, or system logs \cite{zhou-etal-2020-towards},
and employ templates or use humans to rewrite existing utterances for conversation generation.
\citet{DBLP:journals/corr/DodgeGZBCMSW15} and \citet{10.1145/3269206.3271776} employ templates for utterance generation, while \citep{zhou-etal-2020-towards} repurposes existing utterances with human rewrites.
Our method stands apart in using an LM to generate both agent and user utterances.
\citet{lara2022evaluation} emphasize the challenges of evaluating synthetic data from LMs. We address this through both direct crowd-sourced rater evaluation and indirect automated assessment via the performance of LMs that use our synthetic data.

\section{\icer: A Multi-turn User Simulator for Conversational Recommenders}

\label{sec:icer}
% \section{A Behaviorally-grounded \& Multi-turn ConvRec Simulator}

We now detail \icer, a user simulation and synthetic data generation paradigm for CRSs, specifically, those that support \emph{interactive conversational elicitation for recommendation}. Our approach comprises three stages. First, we develop a novel user behavior simulator that interacts with an RS agent that not only recommends items to the user, but also engages in \emph{preference elicitation (PE)} and allows user \emph{item critiques}, which are especially well-suited to NL interaction. We then run sampled synthetic users against the RS to generate two-sided behavior trajectories. In the second stage, we apply \emph{NL templates} to convert each user and RS action into a stylized utterance, similar to other template-based approaches \cite{10.1145/3269206.3271776, DBLP:conf/aaai/RastogiZSGK20}. In the final stage, we use \emph{LM-prompting} with an LM to convert the templatized CRS dialogues into more natural conversations that exhibit the variation one would expect of users.
% engaging with a CRS.
% Figure~\ref{fig:icer_overview} gives an overview of \icer\ framework.

We emphasize that allowable user behaviors are constrained, in part, by the RS/CRS itself, and by those supported by the behavioral simulator. While our CRS agent is fairly general and supports multiple interaction types, the specific \icer\ simulator developed below is somewhat circumscribed. However, the \emph{ICER methodology} is fully general, and can be applied to translate any user behavior simulator into a full-fledged CRS simulator for synthetic data generation and system analysis.

\subsection{The Recommender Agent} % ceb: could join into one subsection with user response models

% Describe the agent actions, mention that if follows a specific policy---give a rough outline (belief state tracker, EVOI, etc.) but not serious details.

Our conversational task is based on a flexible, but non-conversational RS (or agent) providing recommendations from a fixed item corpus $\calI$ (in our movie domain, the set of recommendable movies). The RS can also engage in \emph{preference elicitation (PE)} to query information about the user $u$'s preferences. We consider two types of PE queries: (1) an \emph{item query} $q=S$ presents a slate $S\subseteq\calI$ of $k$ items to $u$ and asks which item is most preferred \citep{ben1985discrete}; (2) an \emph{attribute query} $q=(S,g)$ comprises an item slate $S$ and an attribute (or tag) $g$ and asks $u$ if they would prefer items with more or less of $g$ than the items in $S$ (we restrict $|S| = 1$ in this case). We assume a fixed set of attributes/tags $\calG$ that can be used for attribute queries (e.g., we might ask the user if they want a ``funnier'' or ``less violent'' movie). Recommendations also take the form of a \emph{slate} $S$ of $k$ items, one of which $u$ may \emph{accept}. Should $u$ reject (the items on) slate $S$, they can optionally offer a \emph{critique} of $S$ \citep{chen_critiquing_survey:umuai2012}, specifying an attribute from $\calG$ they would like the recommended item to exhibit more/less of than those in $S$. 

\begin{figure}[]
\centering
\includegraphics[width=.67\textwidth]{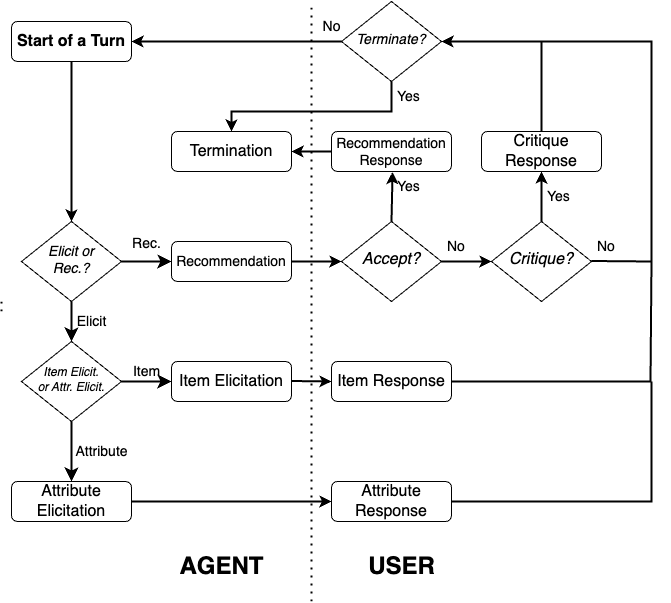}
\caption{The \icer~CRS Simulator Flowchart. The left part depicts decisions (represented with diamonds) and actions (represented with rectangles) in the recommender agent and the right part depicts decisions and actions in user behavior models.}
\label{fig:flowchart}
\end{figure}

At each stage of the interaction with $u$, the RS must choose a recommendation $S$, PE item query $S$, or PE attribute query $(S,g)$ (see Fig.~\ref{fig:flowchart}). In general, the RS can adopt \emph{any} policy (i.e., mapping from history to action choice). In our simulations below, we implement a \emph{specific} RS policy, based in part on the EVOI-based approach of \citet{pe-cav}. The RS maintains a belief state over $u$'s embedding
which is updated in Bayesian fashion given any user action/response.
Let $\calH^{(n)} = \{(q^{(n)},\rho^{(n)})\} \cup\calH^{(n-1)}$, with $\calH^{(0)} = \emptyset$, be the history 
of RS actions $q$ and user responses $\rho$ at turn $n<N$. The RS updates its posterior belief in standard Bayesian fashion:
\begin{equation*} 
P_U(u\! \mid\! \calH^{(n)})\! =\! P(\phi_u\! \mid\! \calH^{(n)})\! \propto P_U(u) \prod_{j=1}^{n} P(\rho^{(j)}\! \mid\! q^{(j)},\phi_u).    
\end{equation*}
This generally does not have a closed form, so we use samples drawn from the unnormalized posterior via Metropolis-Hastings~\citep{chib1995understanding} to update $P_U(u)$.

The RS evaluates potential queries $q$ using
\emph{expected value of information (EVOI)} \citep{preference:aaai02}, which measures the improvement in $u$'s expected utility given their possible responses to $q$. The EVOI of $q$ is
\begin{equation*}
\EVOI(q) = \PEU(q) - \EU^*\big(P_U\big),
\end{equation*}
where $\EU^*(P_U)$ is \emph{maximum expected utility} given belief state $P_U$:
\begin{equation*}
EU^*(P_U) = \max_{i\in \calI} \mathbb{E}_{\phi_u \sim P_U}\big[\phi_u^\top\phi_I(i)\big]
\end{equation*}
% $\calH$ is the current history, 
%
% \begin{equation}
% \label{eq:eu_definition}
% EU^*(P_U(u)) = \max_{i\in \calI} \; \mathbb{E}_{\phi_U(u) \sim P_U(u)}\big[\phi_U(u)^\top\phi_I(i)\big]
% \end{equation}
and $PEU(q)$ is the \emph{posterior expected utility}:
\begin{equation*}
PEU(q) = \mathbb{E}_{\phi_u \sim P_U, \rho\sim P(\rho \mid q, \phi_u)}[EU^*(P_U(u \mid \rho,q))].
\end{equation*}

% \begin{equation}
% \label{eq:peu_definition}
% PEU(q \mid \calH) = \sum_{\rho} P(\rho \mid q,\calH) \cdot EU^*\big(P_U(u \mid \calH\cup\{(\rho,q)\})\big).
% %\max_{i\in S} \mathbb{E}_{\phi_U(u) \sim P_U(u \mid \calH\cup\{(\rho,q)\})}[\phi_U(u)^\top\phi_I(i)]
% \end{equation}
%
A query with maximum PEU maximizes the expected (w.r.t.\ response $\rho$) utility of the user's target item, while EVOI, which is maximized by the same query, measures the \emph{change} in this expected utility induced by the query. 
EVOI-based PE selects the query that offers
% extracts information that offers
maximum expected improvement in recommendation quality.
If the expected utility improvement is non-positive, then the RS may stop since no further information can be obtained about the utility. 
% Query selection
This requires optimizing over the query space, which is linear in the number of attributes but exponential in the number of items.
Under mild assumptions,  PEU can be approximated with the differentiable function:
\begin{equation*}
F(q):=\frac{ \max_{i\in\calI} \|\phi_I(i)\|}{m}\cdot\sum_{\rho} \|\sum_{j=1}^m \phi_{u,j} P(\rho \mid q,\phi_{u,j})\|_2, 
\end{equation*}
where $\{\phi_{u,j}\}_{j=1}^m$ are the $m$ samples drawn from $P_U(u)$ \citep{pe-cav}.
So we instead use gradient-based EVOI optimization \citep{vendrov2020gradient} over this function: $q*\in\arg\max_q F(q)$. 

If the EVOI of the best scoring query does not meet some threshold, the RS instead chooses to make a recommendation $S$ as follows: given a user embedding $\phi_u$ sampled from belief $P_U$, the RS scores all items $i$ in the corpus $\mathcal I$ w.r.t.\ $\hat r_{i,u} = \phi_I(i)^T\phi_u$, and recommends slate $S:=\{i^*_1(u),\ldots,i^*_k(u)\}$ of the $k$ best items.
i.e., $i^*_j(u)\in\!\arg\!\max_{i\in\mathcal I, i\not\in\{i^*_1(u), \ldots,i^*_{j-1}(u)\}} \hat r_{i,u}$.

\subsection{User Behavior Models} % ceb: could join into one subsection with user response models

The first component of the \icer\ simulation framework is a \emph{user behavior simulator} that generates trajectories of user interactions with the RS. Such a simulator should reflect both \emph{within-trajectory behavioral consistency}---user actions at each point in the interaction should be plausible, both individually and jointly---and \emph{distributional consistency}---the distribution of trajectories should roughly correspond to those expected in the user population of interest. Our simulator achieves consistency in a standard way by (a) sampling user embeddings (representing a user's preferences) from a CF model trained on ratings data (see Sec.~\ref{sec:prelim}); and (b) ensuring each user interaction is conditioned on the CRS history and the user's embedding. 

We generate the following user actions (see Fig.~\ref{fig:flowchart}), each dependent on current RS turn (or context) and driven by a specific embedding-conditioned behavioral model, where the user $u$ is represented with embedding $\phi_U(u)$.

\vspace*{1mm}
\noindent
\textbf{Recommendation Acceptance.} When the RS makes a slate recommendation $S$, $u$ evaluates the utility of each $j\in S$, and makes a choice using a standard \emph{multinomial logit} choice model \cite{ben1985discrete}, selecting item $i$ with probability:
% \begin{small}
\begin{equation*}
P(\rho = i \mid S, \phi_U(u)) = \frac{\exp(\phi_I(i)^\top\phi_U(u)/T)}{\sum_{j\in S} \exp(\phi_I(j)^\top\phi_U(u)/T)},
\end{equation*}
where $T$ is a temperature parameter. Rejection of all items in $S$ is captured by including an implicit null item.

\vspace*{1mm}
\noindent
\textbf{Item Query Response.} If during a PE step, $u$ is asked to reveal their favorite from a slate $S$ (an item query), they will noisily respond as if asked to select a recommended item above (but with no null item assumed).

\vspace*{1mm}
\noindent
\textbf{Attribute Query Response.} If an attribute query $q=(S,g)$ is posed, $u$ responds according to the following model \citep{pe-cav}: $u$ compares their \emph{target item} $\phi^*_{I,u}=(\max_{i} \|\phi_I(i)\|_2) \times
\phi_U(u) / \|\phi_U(u)\|_2$\footnote{The \emph{target} item is defined as $\phi^*_{I,u} \in \arg\max_{\phi\in \Gamma} \phi_U(u)^\top \phi$, in which the constrained space $\Gamma:=\{\phi: \|\phi\|_2\leq \max_{i\in\mathcal I} \|\phi_I(i)\|_2\}$, connects the user's ideal item with the familiarity of the item space $\calI$.}
with the mean embedding $\phi_{I,\overline S}=\frac{1}{|S|}\sum_{i\in S} \phi_I(i)$ of items in $S$.
% resulting $\phi^*_{I,u} = \max_{i\in\mathcal I} \|\phi_I(i)\|_2 \cdot \phi_U(u) / \|\phi_U(u)\|_2$.
If the target exhibits more of attribute $g$ than the mean slate, i.e., $c_g^\top\phi^*_{I,u} > c_g^\top\phi_{I,\overline S}$,
$u$ (stochastically) requests \emph{more} of $g$ (response $\rho=+1$) according to a \emph{probit} model \citep{train2009discrete,chaptini2005use}:
\begin{equation*}
P(\rho= +1 \mid q, \phi_U(u)) = \mathbf{\Phi}(c_g^\top(\phi^*_{I,u} - \phi_{I,\overline S})/\sigma_g),
\end{equation*}
where $\mathbf{\Phi}(\cdot)$ is the standard Gaussian CDF, and $\sigma_g$ is the standard-deviation parameter.
% Replacing the user embedding $\Phi_U(u)$ with the sampled belief $\phi_u\sim P_U(u)$, the RS models user responses to queries with $P(\rho = +1 \mid q, \phi_u)$.

\vspace*{1mm}
\noindent
\textbf{Item Critiques.} A user $u$ can also spontaneously \emph{critique} the RS's recommendation $S$ to direct the RS toward more preferred items \cite{chen_critiquing_survey:umuai2012} in cases where $S$ is rejected. Since we use soft attributes whose semantics is given by CAVs (see Sec.~\ref{sec:prelim}), we adopt the spontaneous critiquing model of \citet{gopfert2021discovering}: $u$ chooses to critique the attribute $g$ with maximal \emph{salience}, which (roughly) measures the difference between the average value of $g$ in $S$ and the value of $g$ exhibited by $u$'s target (or \emph{ideal}) item $\phi^*_{u}$. The direction of the critique (more/less) is determined by the attribute query response model above.

% Otherwise, $u$ critiques slate $S$ using the \emph{most salient tag} $g$ with respect to its utility
% \todoch{Can just refer to the CAV paper for definitions of ideal item, most salient tag, and utility improvement.}
% improvement~\cite{gopfert2021discovering}, i.e., $u$ critiques using tag $g^*$, which maximizes the salient score $\delta_u^\top c_g$, where $\delta_u=(\phi^*_{I,u}-\frac{1}{|S|}\sum_{i\in S}\phi_I(i))\odot \phi_U(u)$
% is the utility difference vector between the ideal item $\phi^*_{I,u}$ and average item over slate $S$. 

\vspace*{1mm}
\noindent
\textbf{Other Behaviors.} Our behavior simulator also allows $u$ to terminate stochastically prior to accepting a recommendation, with probability increasing with conversation length. We do not do so here, but note that a host of other behaviors could be modeled
% readily, some of 
which depend on latent user factors, just as the behaviors above depend on latent user preferences (embeddings). For example, session termination could arise due to dissatisfaction (e.g, if recommendation quality does not improve across turns) or frustration (e.g., too many PE queries), possibly in ways that vary across users (e.g., due to the varying patience of different users).

\subsection{Templatized Dialogue Generation}

Given a set of interaction trajectories generated by the behavior simulator (and RS agent), we turn them into \emph{conversational/NL dialogues} representative of a CRS.
% Our first step in doing so is to 
We first apply 
% use templates, with 
a specific template for each user or agent action type.
Such templatization is common for CRS data generation \cite{10.1145/3269206.3271776, DBLP:conf/aaai/RastogiZSGK20}. Fig. ~\ref{fig:templ_dialog} shows an example of templatized dialogue.

\begin{figure}[t]
\begin{tabular}{|p{6in}|}
\hline
\textbf{Agent:} What do you think about My Best Friend's Wedding (1997)? Do you want something more romantic than this? \\
\textbf{User:} No, I want something less romantic. \\
\textbf{Agent:} These are 2 movies you might like: The Basketball Diaries (1995), Hot Shots! (1991) \\
\textbf{User:} No. I don't like them. Do you have something less serious than The Basketball Diaries (1995)? \\
\textbf{Agent:} Which of these movies do you prefer? Beverly Hills Cop II (1987), Take the Money and Run (1969) \\
\textbf{User:} I'd choose Beverly Hills Cop II (1987). \\
\textbf{Agent:} These are 2 movies you might like: The Revenant (2015), How the Grinch Stole Christmas! (1966) \\
\textbf{User:} How the Grinch Stole Christmas! (1966) is what I am looking for! Thanks. \\ \hline
\end{tabular}
\caption{An Example of Templatized Dialogue.}
\vspace{-0.1in}
\label{fig:templ_dialog}
\end{figure}

\subsection{LM-prompted Dialogue Generation}

\begin{figure}[t]
\begin{tabular}{|p{6in}|}
\hline
\textbf{Agent:} Regarding My Best Friend's Wedding (1997), do you prefer a movie with a more romantic storyline? \\
\textbf{User:} I'd prefer something less romantic as I'm not in the mood for love stories. \\
\textbf{Agent:} You may enjoy these non-romantic options: The Basketball Diaries (1995), a gritty drama about drug addiction, or Hot Shots! (1991), a hilarious military satire. \\
\textbf{User:} I'm looking for something lighthearted, as I'm not interested in intense dramas like The Basketball Diaries (1995). \\
\textbf{Agent:} How about these comedies: "Beverly Hills Cop II (1987)", an action-packed crime comedy, or "Take the Money and Run (1969)", a classic mockumentary? Which one do you prefer? \\
\textbf{User:} I'll go with Beverly Hills Cop II (1987) because I'm looking for a lighthearted action comedy. \\
\textbf{Agent:} You may enjoy "The Revenant (2015)", a gripping survival drama, or "How the Grinch Stole Christmas! (1966)", a heartwarming animated classic. \\
\textbf{User:} I'll go with "How the Grinch Stole Christmas! (1966)" as it's a lighthearted classic that fits my current mood. \\ \hline
\end{tabular}
\caption{An Example of LM-prompted Dialogue given Fig.~\ref{fig:templ_dialog}.}
\label{fig:inpaint_dialog}
\end{figure}

Templatized dialogues tend to be repetitive, monotonous and unnatural, and are often restricted to the use of particular key-phrases,
% (e.g., simple ``yes/no''),
which limits their richness.
To overcome this, inspired by \emph{dialogue inpainting} \citep{leszczynski2023}, we use an LM to inject natural variation into templatized dialogues. Fig. ~\ref{fig:inpaint_dialog} illustrates the result of this process.

We first use a prompted LM to replace the simulated user's templatized utterances turn by turn: the LM is given a prompt that includes all previous (LM-modified) user turns, all previous (templatized) RS agent turns, and the current \emph{templatized} user turn, and is asked to rephrase the templatized user turn. This context helps the LM craft a user utterance consistent with her/his generated actions/behaviors. The agent's templatized utterances are then replaced in a similar fashion (using the LM-modified user turns). Figures~\ref{fig:agent_turn_prompt}, \ref{fig:agent_turn_prompt_item}, \ref{fig:agent_turn_prompt_attr} and \ref{fig:user_turn_prompt} show the prompt for each user or agent action type in the movie domain.

While inspired by inpainting, our approach differs slightly, since we do not include \emph{future} templatized turns in the prompt for the current turn. We do so for two reasons: (a) we find the LM sometimes anticipates future information, resulting in unnatural conversations where, say, a user's response seems to anticipate something not yet mentioned by the RS agent; and (b) we use more powerful LM models (see the following section on MD-DICER for details) than were available to \citet{leszczynski2023}, which seems to obviate the need to inject this additional ``forward-looking'' structure.

\begin{figure}[t]
\begin{tabular}{|p{6in}|}
\hline
Agent: ( inpainted utterance ) \\
User: ( inpainted utterance ) \\
% Agent: ( inpainted utterance ) \\
% User: ( inpainted utterance ) \\
... \\
User: ( inpainted utterance ) \\
\textbf{Agent:} ( templatized utterance ) \\
\ \\
Above is a conversation between an agent and a user in turns. The agent tries to find the user's preference by asking questions and then recommends movies the user would like to watch. \\
Rephrase the last Agent turn and it should satisfy following requirements. \\
1. Begin with "Agent:" without new lines. \\
2. Must include the movie title followed by the released year (e.g., Gravity (2013)). \\
3. Include short comments about movies. \\ \hline
\end{tabular}
\caption{Prompt for Recommendation Agent Turn.}
\label{fig:agent_turn_prompt}
\end{figure}

\begin{figure}[t]
\begin{tabular}{|p{6in}|}
\hline
Agent: ( inpainted utterance ) \\
User: ( inpainted utterance ) \\
% Agent: ( inpainted utterance ) \\
% User: ( inpainted utterance ) \\
... \\
User: ( inpainted utterance ) \\
\textbf{Agent:} ( templatized utterance ) \\
\ \\
Above is a conversation between an agent and a user in turns. The agent tries to find the user's preference by asking questions and then recommends movies the user would like to watch. \\
Rephrase the last Agent turn and it should satisfy following requirements. \\
1. Begin with "Agent:" without new lines. \\
2. Must include the movie title followed by the released year (e.g., Gravity (2013)). \\
3. Include short comments about movies. \\
4. Do not recommend the movies but ask preference between them. \\
5. Ask a comparison question at the end. \\ \hline
\end{tabular}
\caption{Prompt for Item Elicitation Agent Turn.}
\label{fig:agent_turn_prompt_item}
\end{figure}

\begin{figure}[t]
\begin{tabular}{|p{6in}|}
\hline
Agent: ( inpainted utterance ) \\
User: ( inpainted utterance ) \\
% Agent: ( inpainted utterance ) \\
% User: ( inpainted utterance ) \\
... \\
User: ( inpainted utterance ) \\
\textbf{Agent:} ( templatized utterance ) \\
\ \\
Above is a conversation between an agent and a user in turns. The agent tries to find the user's preference by asking questions and then recommends movies the user would like to watch. \\
Rephrase the last Agent turn and it should satisfy following requirements. \\
1. Begin with "Agent:" without new lines. \\
2. Must include the movie title followed by the released year (e.g., Gravity (2013)). \\
3. Include short comments about movies. \\
4. Do not recommend the movies. \\ \hline
\end{tabular}
\caption{Prompt for Attribute Elicitation Agent Turn.}
\label{fig:agent_turn_prompt_attr}
\end{figure}

\begin{figure}[t]
\begin{tabular}{|p{6in}|}
\hline
Agent: ( inpainted utterance ) \\
User: ( inpainted utterance ) \\
% Agent: ( inpainted utterance ) \\
% User: ( inpainted utterance ) \\
... \\
Agent: ( inpainted utterance ) \\
\textbf{User:} ( templatized utterance ) \\
\ \\
Above is a conversation between an agent and a user in turns. The agent tries to find the user's preference by asking questions and then recommends movies the user would like to watch. \\
Rephrase the last User turn and it should satisfy following requirements. \\
1. Begin with "User:" without new lines. \\
2. Must be consistent with what the user said in the earlier turns in the conversation. \\
3. Explain very briefly the rationale of the choice. \\ \hline
\end{tabular}
\caption{User Turn Prompt.}
\label{fig:user_turn_prompt}
\end{figure}

\section{MD-DICER: A MovieLens Dialogue Dataset using \icer}
\label{sec:icer_movie}

To generate realistic multi-turn recommendation conversations in the movie domain, we apply the \icer\  methodology using the MovieLens 25M dataset \cite{harper16:movielens}, with $25$ million ratings of $62\,000$ movies by $162\,000$ users. As real users typically only provide informative feedback on familiar movies and our focus is on user uncertainty and PE (and not item cold-start), we focus on a subset of $2\,109$ popular movies with at least $1\,000$ ratings (these movie embeddings are considered extremely stable).
% To ensure users familiarity with the items, we focus on a subset of $2\,109$ popular movies with at least $1\,000$ ratings, allowing for more meaningful elicitation feedback during the data generation process.
We first train $d$-dimensional CF embeddings ($d=128$) to represent users and items, and $d$-dimensional Gaussian user priors $P_U(u)$, with $u$ embedding as the mean and a learnable variance.
% , avoiding extra complexity of modeling the full user covariances.
We also use the human assessment of movie attributes (tags) from the Soft-Attributes dataset~\cite{sigir21:filipandkristian} to learn attribute CAV vectors, following~\citet{cav_journal}. We use the $33$ most informative attributes for critiquing and elicitation.
Our RS simulation treats $u$'s embedding $\phi_U(u)$ as \emph{ground-truth} latent state from which responses are generated. For this reason, we restrict simulation to the $3650$ users, who have rated at least $50$ movies, increasing confidence that their embeddings reflect their underlying preferences \cite{pe-cav}. Note that while our response models have access to this ground-truth embedding, the RS does not---it is given only the prior $P_U(u)$ before its interactions with $u$.

% To ensure the trajectories generated from the \icer~framework is realistic, we focus on interactions from the \emph{power users}, who rate at least $50$ movies. \icer~simulates $100\,000$ trajectories of user-agent movie-related interactions for up to $7$ turns, involving $3\,650$ unique users. The MovieLens Dialogues dataset is then created by inpainting these trajectories into conversations.
% Beyond the dialogues themselves, this dataset also includes MovieLens metadata, e.g., user IDs, probabilities associated with the simulator's response models, Gaussian user priors, and user/item CF embeddings, whose the underlying behaviors align with the conversations.

Our MovieLens Dialogues Dataset, MD-DICER, is generated by first sampling 100,000 user embeddings and generating interactions of up to $7$ agent-user turns for each; second, using templates to convert these into stylized conversations; and third, using an LM to translate these into more natural conversations.
Figures~\ref{fig:templ_dialog} and~\ref{fig:inpaint_dialog} show one templatized dialogue and its LM-prompted counterpart. 
We employ Gemini Ultra \cite{team2023gemini} as our LM, though any suitable off-the-shelf LM could suffice.
Apart from the dialogues, MD-DICER also includes MovieLens and RS metadata 
(e.g., user IDs, embeddings; item embeddings; RS response-model probabilities; etc.).

\section{Evaluation}\label{sec:experiments}

We assess the MD-DICER dataset, and 
% the effectiveness of
our \icer\ methodology, 
% MovieLens Dialogues dataset
% generated by \icer~(see Section \ref{sec:icer_movie})
using both human and automated evaluation. We first assess the intrinsic quality of our generated dialogues using human raters to evaluate properties such as fluency, naturalness, consistency, and redundancy. This provides not only quality metrics for MD-DICER, but a labeled CRS dataset which can serve as a valuable resource for fine-tuning LM-based recommenders using, say, reinforcement learning from human feedback
% (RLHF)
\cite{christiano2017deep,ouyang2022rlhf}. We then evaluate the performance of \icer\ on a downstream recommendation task, showing how prompting an existing LM with \icer-generated dialogues can improve recommendation quality. We intentionally avoid training or fine-tuning our own LM 
% with \icer\ 
to isolate the effectiveness of \icer-dialogues.

\subsection{Rater Evaluation}

For our rater study, we employ a two-stage sampling approach: we first randomly select 600 MovieLens users; then for each user we sample a single dialogue.
% , resulting in a set of 600 synthetic dialogues for an evaluation that spans across eight aspects described below.
Each dialogue is evaluated by three raters along eight dimensions:
% \footnote{The three raters per dialogue are assigned at random; logistical constraints means that full overlap in raters across all dialogues is not achievable. During evaluation, raters rely solely on the dialogue content (no other information, e.g., user ratings history or embeddings, is provided).}

\noindent
\textbf{Progress.} The agent's understanding of the user's preferences is steadily improving with each interaction in the conversation.

\noindent
\textbf{Behavioral Consistency.} A user's expressed preferences in response to RS queries are aligned (i.e., they reflect the preferences of a single, coherent individual) across turns.
    % The preferences of the user are consistent, which means choices or responses to the agent question / recommendations seem to come from the same person under similar contexts. User’s response in a turn is consistent with other responses in earlier turns in the same conversation.

\vspace*{1mm}
\noindent
\textbf{Non-redundancy.} The system avoids repetitive interactions, e.g., asking redundant questions or repeating similar recommendations.

\vspace*{1mm}
\noindent
\textbf{Factuality.} User-agent interactions are based on accurate movie information.

\vspace*{1mm}
\noindent
\textbf{Context Consistency.} The conversation maintains continuity by ensuring all responses from both the user and the agent are relevant, coherent, and stay on topic.

\vspace*{1mm}
\noindent
\textbf{Attribute Consistency.}  The agent coherently understands the user's preferences (w.r.t. attributes) in the conversation and makes tailored recommendations.

\vspace*{1mm}
\noindent
\textbf{Naturalness.} The conversation is smooth and engaging, with relevant responses that flow seamlessly between turns.

\vspace*{1mm}
\noindent
\textbf{Fluency.} Agent and user utterances exhibit linguistic and semantic fluency.

% \begin{itemize}
%     \item \textbf{Progress.} The agent's understanding of the user's preferences is steadily improving with each interaction in the conversation.
%     \item \textbf{Behavioral Consistency.} The alignment of a user's expressed preferences in responses to various agent's queries. This notion indicates that the user's choices reflect the preferences of a single, coherent individual across similar contexts.
%     % The preferences of the user are consistent, which means choices or responses to the agent question / recommendations seem to come from the same person under similar contexts. User’s response in a turn is consistent with other responses in earlier turns in the same conversation.
%     \item \textbf{Non-redundancy.} The system avoids repetitive interactions, e.g., asking redundant questions or suggesting similar recommendations repeatedly. 
%     \item \textbf{Factuality.} User-agent interactions are based on accurate movie information.
%     \item \textbf{Context Consistency.} Conversation maintains continuity by ensuring all responses from both the user and the agent are relevant, coherent, and stay on topic.
%     \item \textbf{Attribute Consistency.}  The agent coherently understands user's preferences (w.r.t. attribute tags) in the conversation and makes tailored recommendations.
%     \item \textbf{Naturalness.} Conversation is smooth and engaging, with relevant responses that flow seamlessly between turns.
%     \item \textbf{Fluency.} Agent and user utterances exhibit linguistic and semantic fluency. 
% \end{itemize}

Raters engage in a two-step evaluation. First, they answer yes-no questions to determine if \icer-generated dialogues exhibit the characteristics in question. If they answer ``no,'' they are then asked to specify problematic turns and provide detailed, open-ended explanations. %\footnote{Fig. 10 (Appendix A) shows the rater form used; Fig. 11 shows the one for pairwise comparisons, discussed next.} 
This yields both dialogue-level labels (yes/no) and turn-level annotations
% for supervised model training, allowing the model to learn from positive and negative examples at both the conversation and turn-level. Fig.~\ref{fig:sample_form} in App.~\ref{app:eval} shows the form used for evaluation.
that can be used for supervised model training or fine-tuning.
Table~\ref{tab:human_eval} shows that the fraction of positive rater feedback for each aspect is very high (close to 100\%) with low variance, suggesting that \icer-generated dialogues are compelling examples of multi-turn CRS interactions.

% and its 95\% confidence intervals, which are all close to $100\%$, demonstrating the intrinsic quality of inpainted dialogues. 
% Discussion 

\begin{table}[t]
% \begin{minipage}[t]{0.48\textwidth}
\centering
    \begin{tabular}{||c|c||} \hline
        Progress & $94.6 \pm 1.5$\% \\ 
        Behavioral Consistency & $97.8 \pm 0.7$\% \\
        Non-redundancy & $97.4 \pm 0.8$\% \\
        Factuality & $98.1 \pm 0.6$\% \\
        Context Consistency & $98.9 \pm 0.4$\% \\
        Attribute Consistency & $98.7 \pm 0.5$\% \\
        Naturalness & $99.1 \pm 0.4$\% \\
        Fluency & $99.4 \pm 0.3$\% \\ \hline
    \end{tabular}
% \end{minipage}
\caption{Fraction of positive rater feedback for each evaluation criterion (with 95\% Confidence Intervals.).}
\vspace{-0.1in}
\label{tab:human_eval}
\end{table}

To assess the impact of \icer's LM-based prompting on dialogue quality, we next ask raters to compare original templatized conversations with their LM-prompted counterparts. For each of 600 conversation pairs, three raters compare them along the eight dimensions above, and on overall quality (is templatized better/worse than or similar to prompted). Raters are not made aware of which dialogue is prompted or templatized (though the prompted variants are usually longer), nor do they
% Unlike the assessment above, raters do not
provide turn-based annotations or open-ended feedback. 
% CEB: Keep as one para
% For all 600 inpainted dialogues, we also did a pairwise comparison in which three raters which might be different from ones in the previous evaluation/annotation compare an inpainted dialogue with the corresponding templatized version. Expressing preferences over different LM answers is easier for humans to annotate labels for reward model training in RLHF~\cite{ziegler20}. We asked which one among two dialogues is better in terms of the aspect being evaluated or they are about the same. We did not disclose whether a dialogue is an inpainted one or a templatized one to raters although the inpainted version is usually longer. We did not require raters to provide any turn-level annotations nor open-ended feedback. In addition to the eight aspects, we also asked overall which one is better between an inpainted dialogue and its templatized one.
%
Table~\ref{tab:pairwise_comparison} summarizes the rater comparisons across the eight dimensions and overall conversation quality. The LM-prompted dialogues are viewed to have higher quality than their templatized counterparts over $80\%$ of the time. Similarly, prompted dialogues are favored significantly for six of the  eight aspects. While prompting preserves recommendation and query quality, the LM sometimes introduces background information that is deemed redundant, impacting \emph{non-redundancy}.
% Rater scores for "Naturalness" and "Fluency" are also lower for inpainted dialogues. This suggests that raters may not expect user utterances to be very elaborate, and may have found the templatized dialogues sufficiently fluent in some cases.
Regarding \emph{fluency},
% shows some preference for prompted dialogues,
prompted and templatized are score equally well, in part due to the fact that the templates themselves are ``sensible.''
% The over-elaboration (see \emph{Redundancy}) generated by LMs may also play a role.

\begin{table}[t]
    \centering
    \begin{tabular}{||@{}c@{}||c|c|c||} \hline
        Aspect & Similar & Prompted & Template \\ \hline\hline
        Overall & 7.9 $\pm$ 1.1 \% & 84.1 $\pm$ 1.5\% & 8.0 $\pm$ 1.1\% \\ 
        Progress & 20.5 $\pm$ 1.6 \% & 71.4 $\pm$ 1.8 \% & 8.1 $\pm$ 1.1 \% \\
        Behavioral Consistency & 23.3 $\pm$ 1.7 \% & 67.9 $\pm$ 1.9 \% & 8.8 $\pm$ 1.2 \% \\
        Non-redundancy & 35.1 $\pm$ 1.9 \% & 37.1 $\pm$ 2.0 \% & 27.8 $\pm$ 1.8 \% \\
        Factuality & 19.2 $\pm$ 1.6 \% & 73.1 $\pm$ 1.8 \% & 7.7 $\pm$ 1.1 \% \\
        Context Consistency & 31.9 $\pm$ 1.9 \% & 61.1 $\pm$ 2.0 \% & 7.0 $\pm$ 1.0 \% \\
        Attribute Consistency & 17.8 $\pm$ 1.6 \% & 72.7 $\pm$ 1.8 \% & 9.5 $\pm$ 1.2 \% \\
        Naturalness & 23.1 $\pm$ 1.7 \% & 62.1 $\pm$ 2.0 \% & 14.8 $\pm$ 1.4 \% \\
        Fluency & 42.1 $\pm$ 2.0 \% & 44.0 $\pm$ 2.0 \% & 13.9 $\pm$ 1.4 \% \\ \hline
    \end{tabular}
    \caption{Pairwise Comparison of Evaluation Criteria (LM-Prompted vs.\ Templatized with 95\% Confidence Intervals).}
\vspace{-0.1in}
    \label{tab:pairwise_comparison}
\end{table}

% Table~\ref{tab:pairwise_comparison} presents the distribution of three possible choices for each aspect evaluated in the pairwise comparison study. More than 80\% of time raters thought inpainted dialogue is better overall than the corresponding templatized one. The inpainted version also outperformed the templatized version most of time on all eight aspects evaluated. Probably the ``Non-redundancy'' aspect is the worst for inpainted dialogues. Although the inpainting process does not change the recommendations or the gist of questions, it tries to add more background information and possibly add some redundancy. The inpainted version did not get very high score on both ``Naturalness'' and ``Fluency''. Perhaps raters did not expect user utterances to be as wordy as our inpainted dialogues and they thought templatized dialogues are already fluent in some cases.

\subsection{Automated Evaluation}

We now describe an automated evaluation process to assess the benefit of using natural dialogue in a recommendation task, especially in service of NL-based preference elicitation and user critiquing. We use (partial and complete) \icer-generated conversations as a key component of this evaluation.
% \icer~employs an automated evaluation process to assess the benefit of dialogues in a recommendation task. 
Our process leverages the Gemini Pro LM \cite{team2023gemini}, which we use to evaluate recommendations to users
% recommendation slates, from a pre-selected set of candidate items,
given a specific conversation history. Unlike above---where an actual RS makes the recommendations when generating user simulations---here \emph{the LM itself is tasked with making recommendations}, albeit in a limited way. For each user $u$, we assume access to a synthetic dialogue $C_u$, the prior belief $P_U(u)$, $u$'s CF embedding $\phi_U(u)$, and item embeddings $\phi_I(\cdot)$; the LM makes recommendations based on the first two elements, while the last two allow computation of the user's utility for any recommended item. Our goal is to quantify the improvement in the LM's recommendation quality \emph{driven by the inclusion of ICER-generated dialogues as part of the prompt}, compared to an LM that uses only prior knowledge of $u$.

% In automatic evaluation, we evaluate the benefit of dialogues in \icer~on a  recommendation task in which Gemini Pro\cite{team2023gemini} returns a slate from a pre-selected item set $\hat{I}(u)$ described in the prompt. For each MovieLens user $u$ we have their synthetic dialogue $C(u)$ with the agent, their ground-truth embedding $\phi_U(u)$ unknown to the RS and corresponding prior distribution $P_U(u)$ known to the RS. We also have the ground truth movie embeddings $\phi_I(i)$ for $i \in \mathcal I$ during WALS training. We then can evaluate the performance according to $\phi_U(u) ^T \phi_I(i)$ and quantify the benefit added by $C(u)$ given the RS already knows $P_U(u)$. 

\vspace*{1mm}
\noindent
\textbf{Mapping User Embeddings to Text Profiles for Prompting.}
LMs like Gemini typically handle text and multi-modal input---there have yet to emerge well-established methods for them to consume, say, uninterpretable embeddings or distributions of user behaviors, as would be needed to ingest our priors (though see, e.g., \citet{tennenholtz2024demystifying,li-liang-2021-prefix}). 
% \footnote{ADD A COUPLE CITES TO PAPERS THAT TRY TO DO THIS: LIANG, DEMYSTIFYING, ETC.}
We address this by converting user preference embeddings into textual descriptions as follows.
% While LMs like Gemini typically process text input, they cannot directly handle numerical embeddings or distributions of these embeddings. To bridge this gap, we transform user preference embeddings into textual descriptions.
We sample users with a large number of ratings as above. For a sampled user $u$, we generate a heuristic \emph{text profile} that (roughly) mimics the RS's prior $P_U(u)$ over $u$'s preferences, including its uncertainty. We restrict profiles to a vocabulary $V_\textrm{prof}$ of popular movies. We generate a profile for $u$ as follows: (1) We first sample $M$ embeddings $\phi^m_u\sim P_U(u), m\leq M$ from $u$'s prior. (2) For each $\phi^m_u$, we compute the cosine similarity of each item $i\in V_\textrm{prof}$, which gives an empirical distribution $H_u(i)$ over $u$'s score for $i$. (3) We select \emph{liked} items $L_u$,  \emph{disliked} items $D_u$, and \emph{uncertain} items $N_u$ for $u$'s profile, with $|L_u|\!+\! |D_u|\! +\! |N_u|\! =\! K$: liked items have a high mean score w.r.t.\ $H_u(i)$ with reasonably low variance; disliked items are similar with low score; and uncertain items have neither high nor low mean, and a distribution $H_u(i)$ with high variance. (4) Along with (partial) in-painted dialogues $C_u$, which often include context (e.g., reasoning for a user response), we then convert this into a text profile $F_u$  of user preferences using few-shot, chain-of-thought prompting \cite{chain_of_thought}.\footnote{Specifically, we adopt a ``first-person'' LM prompt, asking it to play the role of a user, and priming its preferences as: \emph{Q: Do you like movies $L_u$? A: Definitely yes. Q: Do you like movies $N_u$? A: I am not sure as I have not watched them.}}

\begin{table*}[t]
    \centering
    \begin{tabular}{||c||c|p{1.5in}|p{1.5in}||}\hline
       Num.\ Turns & MD-DICER & Dual-agent LM \textbf{with} ICER-style PE prompt & Dual-agent LM \textbf{without} ICER-style PE prompt\\ \hline\hline
       0 & \multicolumn{3}{c||}{$0.515 \pm 0.016$} \\
       1 & $0.574 \pm 0.016$ & $0.562 \pm 0.016$ & $0.531 \pm 0.016$ \\
       2 & $0.583 \pm 0.016$ & $0.582 \pm 0.016$ & $0.587 \pm 0.016$ \\
       3 & $0.590 \pm 0.016$ & $0.586 \pm 0.016$ & $0.602 \pm 0.016$ \\
       4 & $0.587 \pm 0.016$ & $0.588 \pm 0.016$ & $0.587 \pm 0.017$ \\
       5 & $0.592 \pm 0.017$ & $0.591 \pm 0.017$ & $0.578 \pm 0.017$ \\
       6 & $0.604 \pm 0.023$ & $0.596 \pm 0.023$ & $0.567 \pm 0.024$ \\
       7 & $0.613 \pm 0.023$ & $0.597 \pm 0.024$ & $0.566 \pm 0.024$ \\ \hline
    \end{tabular}
    \caption{Accuracy in Automatic Evaluation (with 95\% Confidence Intervals.).}
\vspace{-0.1in}
    \label{tab:auto_eval_accuracy}
\end{table*}

\begin{table*}[h]
    \centering
    \begin{tabular}{||c||c|p{1.5in}|p{1.5in}||}\hline
       Num.\ Turns & MD-DICER & Dual-agent LM \textbf{with} ICER-style PE prompt & Dual-agent LM \textbf{without} ICER-style PE prompt\\ \hline\hline
      0 & \multicolumn{3}{c||}{$0.744 \pm 0.009$} \\
      1 & $0.774 \pm 0.009$ & $0.767 \pm 0.009$ & $0.752 \pm 0.009$ \\
      2 & $0.779 \pm 0.009$ & $0.778 \pm 0.009$ & $0.781 \pm 0.009$ \\
      3 & $0.783 \pm 0.009$ & $0.779 \pm 0.009$ & $0.789 \pm 0.009$ \\
      4 & $0.781 \pm 0.009$ & $0.780 \pm 0.009$ & $0.780 \pm 0.009$ \\
      5 & $0.784 \pm 0.009$ & $0.782 \pm 0.009$ & $0.777 \pm 0.009$ \\
      6 & $0.788 \pm 0.013$ & $0.785 \pm 0.013$ & $0.771 \pm 0.013$ \\
      7 & $0.793 \pm 0.013$ & $0.786 \pm 0.013$ & $0.771 \pm 0.013$ \\ \hline
    \end{tabular}
    \caption{NDCG in Automatic Evaluation (with 95\% Confidence Intervals.).}
\vspace{-0.1in}
    \label{tab:auto_eval_ndcg}
\end{table*}

\vspace*{1mm}
\noindent
\textbf{Dual-agent LM as a Baseline.}
We also assess the value of the MD-DICER dataset by comparing it with datasets generated \emph{without using the ICER methodology}. An obvious baseline is to have one LM playing the role of the recommendation agent and another LM playing the role of the user. We use this \emph{dual-agent framework} to generate a dialogue dataset. We assume both the recommendation agent and the user have access to prior user behavioral data in the form of text profile $F_u$.
The recommendation agent initiates a seven-turn dialogue with the information in $F_u$. At turn $n$, we prompt the LM with both $F_u$ and the dialogue so far $C_u^{(n)}$. We use two variants of the prompting strategy. The first is the \emph{dual-agent LM with ICER-style PE prompting} (see Fig.~\ref{fig:dual_agent_with_icer}). We constrain the dialogue with the ICER preference elicitation APIs by instructing the recommender LM to use three options for actions: item elicitation, attribute elicitation, and recommendation, along with an example dialogue from MD-DICER. For the user LM, we also prompt to allow for the possible critiquing of recommendations. The second strategy is the \emph{Dual-agent LM without ICER-style PE prompting} (see Fig.~\ref{fig:dual_agent_without_icer}). In this case, we give the LM maximal flexibility and simply ask the recommendation agent to make recommendations at least once during the seven-turn dialogue. As in MD-DICER, we employ Gemini Ultra \cite{team2023gemini} as our LM to generate baseline dialogue datasets.

\begin{figure}[t]
\small
\begin{tabular}{|p{6in}|}
\hline
Q: Would you like action movies such as Lethal Weapon 2 (1989) or more emotional movies such as The Fault in Our Stars (2014)?\\
A: I'd prefer something more emotional, like The Fault in Our Stars (2014).\\
Q: Would you prefer movies with a strong female lead like Antonia's Line (Antonia) (1995) or movies with a more lighthearted tone like About Time (2013)?\\
A: I'd prefer movies with a more lighthearted tone like About Time (2013).\\
Q: Would you prefer a movie that's more thought-provoking, like The Theory of Everything (2014), or more escapist, like About Time (2013)?\\
A: I'd prefer a movie that's more escapist, like About Time (2013).\\
Q: Considering your preference for lighthearted and escapist movies, would you like to explore movies with a comedic or fantasy element?\\
A: I'd like to explore movies with a comedic element.\\
Q: Based on your preferences so far, would you prefer comedies with a romantic twist like Love Actually (2003) or more slapstick comedies like Airplane! (1980)?\\
A: I'd prefer comedies with a romantic twist like Love Actually (2003).\\ \hline
% Q: Regarding the movies you mentioned, would you prefer a more action-packed movie like Lethal Weapon 2 (1989) or a more heartwarming movie like The Perks of Being a Wallflower (2012)?\\
% A: I'd prefer a more heartwarming movie like The Perks of Being a Wallflower (2012).\\
% Q: Based on the movies you mentioned, it seems you might enjoy movies with heartwarming and emotional themes. Would you prefer a movie that focuses on family relationships, like Little Miss Sunshine (2006), or one that explores coming-of-age experiences, like The Breakfast Club (1985)?\\
% A: I'd prefer a movie that explores coming-of-age experiences, like The Breakfast Club (1985).
\end{tabular}
\caption{Dual-agent LM with ICER-style PE prompt. Example of generated conversation.}
\label{fig:dual_agent_with_icer}
\end{figure}

\begin{figure}[t]
\small
\begin{tabular}{|p{6in}|}
\hline
Q: **Assistant:** Hi there, movie lover! It looks like you're exploring some interesting titles. Is there anything specific you're looking for in a movie today? Action, adventure, animation, or something else entirely?\\
A: **User:** I'm not sure exactly. I like action and adventure, but I'm also open to animation if it's well-made and has a good story.\\
Q: **Assistant:** Great! How about we explore those genres further? Do you prefer fast-paced action, epic battles, or more character-driven adventures? And for animation, are you interested in classic Disney-style films or something more unique and experimental?\\
A: **User:** For action, I enjoy epic battles and fast-paced action. As for animation, I'm interested in both classic Disney-style films and more unique and experimental ones, as long as the story is engaging.\\
Q: **Assistant:** Fantastic! Based on your preferences, would you be interested in movies with thrilling spaceship battles, imaginative animated worlds, or a combination of both?\\
A: I would be interested in a combination of both thrilling spaceship battles and imaginative animated worlds.\\
Q: **Assistant:** Sounds great! Would you prefer a movie with a more serious tone or something with plenty of humor? And do you have any favorite actors or directors whose work you enjoy?\\
A: I would prefer a movie with a balance of seriousness and humor. I don't have any particular favorite actors or directors, but I appreciate well-developed characters and engaging storylines.\\
Q: **Assistant:** Wonderful! I think we're getting closer to finding your perfect movie. Would you like to explore movies that feature diverse and compelling characters, or are you more interested in movies with intricate and mind-bending plots?\\
A: I would like to explore movies that feature diverse and compelling characters.\\ \hline
% Q: **Assistant:** Excellent! I have a selection of movies that might suit your taste. Would you like to hear about a movie with a diverse cast of characters engaged in thrilling spaceship battles, or an animated film featuring unique characters in an imaginative world?\\
% A: I would like to hear about a movie with a diverse cast of characters engaged in thrilling spaceship battles.\\
% Q: **Assistant:** Great choice! How about "Guardians of the Galaxy" (2014)? It features a diverse cast of characters, including a human, a raccoon, and a tree, who team up to save the galaxy in epic spaceship battles. Would you like to hear more about this movie?\\
% A: Yes, I would like to hear more about "Guardians of the Galaxy" (2014).
\end{tabular}
\caption{Dual-agent LM without ICER-style PE prompt. Example of generated conversation.}
\label{fig:dual_agent_without_icer}
\end{figure}

\vspace*{1mm}
\noindent
\textbf{Evaluation.}
We test the ability of an LM to make (limited) recommendations by: (1) using $u$'s text profile $F_u$ as a set of in-context examples to prompt the LM; and (2) asking the LM to compare a \emph{pair} of items. This is equivalent to making a recommendation $\text{LM}(\hat{\mathcal I}; F_u)$ from a (trivial) item set where $|\mathcal{I}| = 2$. We focus on small sets so we do not overly rely on the LM's retrieval capabilities or ability to handle exceedingly long prompts (this is not our focus).\footnote{Specifically, for an item pair $(i_0, i_1)$, we use a simple prompt \emph{Q: Considering your preference demonstrated above do you like $i_0$ more than $i_1$? Please just answer YES or NO. A:}.}
% $\text{LM}(\hat{\mathcal I};\mathcal{I}^*_u, \mathcal{I}^\#_u)$
We can also ask the LM to make recommendations given a complete \icer-generated dialogue $C_u$ with $u$, or a prefix $C_u^{(n)}$ of its first $n$ turns,
$\text{LM}(\hat{\mathcal I}; F_u | C_u)$ or $\text{LM}(\hat{\mathcal I}; F_u | C_u^{(n)})$, respectively.
We then evaluate the ability of the LM to select the better (more preferred) item, \emph{a priori} or at any point in the conversation, w.r.t.\ $u$'s ground-truth embedding.

We test the LM's conversation-conditioned recommendation performance using all $3650$ MovieLens users in the MD-DICER dataset (with $M=500$ embedding samples used to generate each $u$'s ``uncertain'' profile). 
We use $K=10$ movies in preference profiles, and sample a single conversation (and corresponding ground-truth embedding) for each user from MD-DICER.
% (recall a single user may have multiple conversations in the dataset). 
We generate a random item pair $\hat{\mathcal I}$ for each given conversation by sampling one ``reasonably'' preferred (uniformly from the top quartile w.r.t.\ $u$'s ground-truth scores) and one non-preferred (bottom quartile) movie, avoiding overlap with movies in $u$'s profile. Then, for each conversation turn from $t=0$ (only $F_u$ is used) to $t=7$, we assess $\text{LM}(\hat{\mathcal I};F_u | C_u^{(n)})$, i.e., which item the LM believes is preferred by $u$.

% Since this experiment focuses on the effectiveness of \icer~dialogue in aiding the LM to perform the recommendation task rather than its ability to handle complex tasks with long prompts, we limit the evaluation to single-item recommendations with $|\hat{\mathcal I}| = 2$, i.e., comparing randomly generated pairs of items with a simple prompt like \textbf{Q: Considering your preference demonstrated above do you like $i_0$ more than $i_1$? Please just answer YES or NO. A:}

% Evaluating how well our LM-based recommender $\text{LM}(\hat{\mathcal I};F_u | C_u[t])$ performs with standard RS metrics is crucial.
% Three key metrics for this are NDCG \cite{jarvelin2002cumulated}, precision, and recall. NDCG considers the ranking relevance between the slate returned by the LM and the ground-truth ordering based on CF scores $\phi_U(u)^\top\phi_I(i)$, $\forall i \in\hat{\mathcal I}$. Precision measures the portion of LM-retrieved items that are relevant, while recall focuses on the percentage of relevant items the LM returns. 
% %
% Consequently, with each recommendation containing only one relevant item, precision and recall become equivalent metrics.

Table~\ref{tab:auto_eval_accuracy} and Table~\ref{tab:auto_eval_ndcg} show recommendation \emph{accuracy}---how often the best movie is selected by the LM---and NDCG---treating this as a ranking of the two-item set $\hat{\mathcal I}$, to account for their score differences---across conversation turns. We see that incorporating more turns of \icer\  dialogue allows the LM to rate items in a way that is better aligned with user preferences (see the MD-DICER columns in the tables). Using conversations generated by the Dual-agent LM with ICER-style PE prompting results in performance gains across dialogue turns, mirroring the trend observed with MD-DICER. However, the magnitude of these improvements is, on average, less than that achieved by MD-DICER across all seven turns. By contrast, the Dual-agent LM without ICER-style PE prompting produces more unconstrained conversations in which the occurrence of explicit preference elicitation at a given turn is not guaranteed. Consequently, these free-form dialogues fail to yield the consistent increasing trend in both accuracy and NDCG that is characteristic of both MD-DICER and the Dual-agent LM without ICER-style PE prompting.

We note that the LM recommendation performance is significantly improved by the inclusion of PE interactions within CRSs (powered by EVOI or other query-selection methods). This suggests an opportunity to use datasets like MD-DICER to improve LM-based CRSs to better account for PE, user critiquing and other forms of user interaction. We also note that this simple test is meant merely as an illustration of one possible use case of \icer. Extensions of this assessment are left for future work.

\section{Conclusion}
\label{sec:conclusions}

% We introduce \icer, a novel methodology for generating multi-turn conversational recommendation dialogues. \icer\ leverages existing user-item ratings and user-tag data to create a simulation framework that goes beyond simple recommendations to incorporate user critiques and RS-driven preference elicitation. Coupled with in-painting, this results in richer, and natural, sequential user-RS conversational interactions, as demonstrated by rater evaluation of the MD-DICER dataset. Our evaluation of LM-based CRSs with partial conversations demonstrates one (simple) way of using the dataset---many other uses are possible. We emphasize that the \icer\ methodology supports use cases beyond the specific RS and behavior models used for illustration here, and can be applied to a variety of non-RS conversational tasks for synthetic data generation when real conversations are scarce or not available for training (e.g., due to privacy considerations).

\icer, a new methodology, generates multi-turn conversational recommendations. It uses existing user-item ratings and tags to create a simulation framework beyond simple recommendations, incorporating critiques and preference elicitation. With LM-prompting, this yields richer, natural user-RS interactions, evaluated via the MD-DICER dataset. Our evaluation of LM-based CRSs with partial conversations shows one dataset use; many more are possible. The \icer\ methodology applies beyond these specific RS and behavior models to various non-RS conversational tasks for synthetic data generation when real conversations are scarce.

\section*{Declaration on Generative AI}
  The authors have not employed any Generative AI tools during the preparation of this paper.

 %  {\em Either:}\newline
 %  The author(s) have not employed any Generative AI tools.
 %  \newline
  
 % \noindent{\em Or (by using the activity taxonomy in ceur-ws.org/genai-tax.html):\newline}
 % During the preparation of this work, the author(s) used X-GPT-4 and Gramby in order to: Grammar and spelling check. Further, the author(s) used X-AI-IMG for figures 3 and 4 in order to: Generate images. After using these tool(s)/service(s), the author(s) reviewed and edited the content as needed and take(s) full responsibility for the publication’s content.
 
% \bibliographystyle{ACM-Reference-Format}
\bibliography{long,custom}

\newpage

\appendix
\section{Appendix}

\subsection{ICER Framework Overview}

\begin{figure*}[h]
\centering
\includegraphics[width=0.8\textwidth]{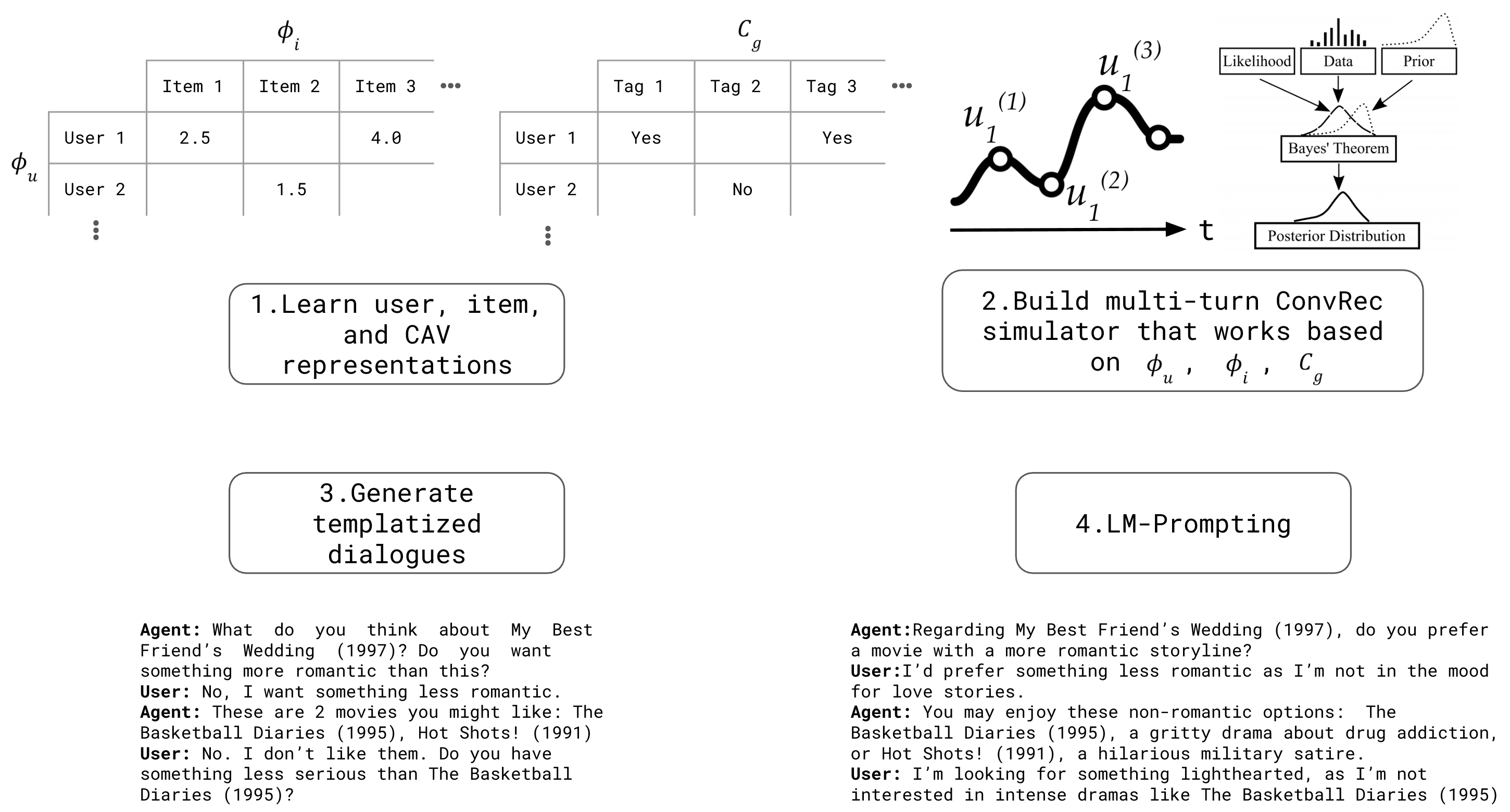}
\caption{Overview of ICER framework. which (i) learns representations of user, item, and CAV using the existing user-item-tag-rating data, (ii) constructs a multi-turn ConvRec simulator, where an agent can query user preferences and recommend items while a user can respond to queries or critique recommendations, (iii) generates dialogue data using templates over the simulator trajectories, and (iv) refines simulated conversations with LM-prompting.}
\label{fig:icer_overview}
\end{figure*}

\subsection{Example Simulation Trajectory}

\begin{lstlisting}[language=json,basicstyle=\small]
user_info {
  id: 19800
  embedding: ...
}
turns {
  agent {
    query {
      mode: ATTR
      slate {
        items {
          id: 1569
          name: "My Best Friend's Wedding (1997)"
        }
      }
      attr {
        id: 23
        direction: 1
        name: "romantic"
      }
    }
  }
  user {
    elicit_resp {
      mode: ATTR
      attr_resp: 0
    }
  }
}
turns {
  agent {
    slate {
      items {
        id: 147
        name: "The Basketball Diaries (1995)"
      }
      items {
        id: 5541
        name: "Hot Shots! (1991)"
      }
    }
  }
  user {
    rec_resp {
      mode: REJECT_FEEDBACK
      attr {
        id: 25
        direction: -1
        name: "serious"
      }
    }
  }
}
turns {
  agent {
    query {
      mode: ITEM
      slate {
        items {
          id: 4084
          name: "Beverly Hills Cop II (1987)"
        }
        items {
          id: 1963
          name: "Take the Money and Run (1969)"
        }
      }
    }
  }
  user {
    elicit_resp {
      mode: ITEM
      item_idx: 0
    }
  }
}
turns {
  agent {
    slate {
      items {
        id: 139385
        name: "The Revenant (2015)"
      }
      items {
        id: 52435
        name: "How the Grinch Stole Christmas! (1966)"
      }
    }
  }
  user {
    rec_resp {
      mode: ACCEPT
      item_idx: 1
    }
  }
}
\end{lstlisting}

\subsection{Details of LM-prompting}
\label{app:inpainting}

\begin{figure*}[h]
\begin{lstlisting}[language=python,basicstyle=\small]
def inpaint_turn(inpainted_dialogue, turn):
    if is_agent_turn(turn):
        inpainted_text = inpaint_agent_turn(inpainted_dialogue, turn)
        return f"\nAgent: {inpainted_text}"
    else:  # user turn
        inpainted_text = inpaint_user_turn(inpainted_dialogue, turn)
        return f"\nUser: {inpainted_text}"

def inpaint_dialogue(templatized_dialogue):
    inpainted_dialogue = ""
    for turn in templatized_dialogue.turns:
        inpainted_dialogue += inpaint_turn(inpainted_dialogue, turn)
    return inpainted_dialogue
\end{lstlisting}
\caption{LM-prompting Algorithm written in Python syntax. $inpaint\_agent\_turn()$ and $inpaint\_user\_turn()$ call LMs with a prompt. They use different prompts and refer to Figures \ref{fig:agent_turn_prompt}, \ref{fig:agent_turn_prompt_item}, \ref{fig:agent_turn_prompt_attr} and \ref{fig:user_turn_prompt} for examples.}
\label{fig:inpainting_alg}
\end{figure*}

\begin{mmbox}{}
\textbf{Agent:} What do you think about My Best Friend's Wedding (1997)? Do you want something more romantic than this? \\
\textbf{User:} No, I want something less romantic. \\
\textbf{Agent:} These are 2 movies you might like: The Basketball Diaries (1995), Hot Shots! (1991) \\
\textbf{User:} No. I don't like them. Do you have something less serious than The Basketball Diaries (1995)? \\
\textbf{Agent:} Which of these movies do you prefer? Beverly Hills Cop II (1987), Take the Money and Run (1969) \\
\textbf{User:} I'd choose Beverly Hills Cop II (1987). \\
\textbf{Agent:} These are 2 movies you might like: The Revenant (2015), How the Grinch Stole Christmas! (1966) \\
\textbf{User:} How the Grinch Stole Christmas! (1966) is what I am looking for! Thanks.
\end{mmbox}

\begin{mmbox}{}
\textbf{Agent:} Which of these movies do you prefer? The Raid: Redemption (2011), Seven Psychopaths (2012) \\
\textbf{User:} I'd choose Seven Psychopaths (2012). \\
\textbf{Agent:} What do you think about White Squall (1996)? Do you want something denser than this? \\
\textbf{User:} No, I want something less dense. \\
\textbf{Agent:} These are 2 movies you might like: Get Shorty (1995), Speed (1994) \\
\textbf{User:} No. I don't like them. Do you have something less cartoonish than Get Shorty (1995)? \\
\textbf{Agent:} These are 2 movies you might like: Blue Velvet (1986), Shadow of a Doubt (1943) \\
\textbf{User:} Blue Velvet (1986) is what I am looking for! Thanks.
\end{mmbox}

\begin{figure*}[h]
\begin{mmbox}{}
\textbf{Agent:} Which of these movies do you prefer? Miracle on 34th Street (1947), Harvey (1950) \\
\textbf{User:} I'd choose Miracle on 34th Street (1947). \\
\textbf{Agent:} These are 2 movies you might like: A Room with a View (1986), Wild Things (1998) \\
\textbf{User:} No. I don't like them. Do you have something funnier than A Room with a View (1986)? \\
\textbf{Agent:} Which of these movies do you prefer? Bad Santa (2003), Wayne's World (1992) \\
\textbf{User:} I'd choose Bad Santa (2003). \\
\textbf{Agent:} These are 2 movies you might like: Sophie's Choice (1982), The Accountant (2016) \\
\textbf{User:} No. I don't like them. Do you have something less dynamic than Sophie's Choice (1982)? \\
\textbf{Agent:} These are 2 movies you might like: Silver Linings Playbook (2012), Mud (2012) \\
\textbf{User:} No. I don't like them. Do you have something with a less unique story than Silver Linings Playbook (2012)? \\
\textbf{Agent:} These are 2 movies you might like: The Grand Budapest Hotel (2014), A Few Good Men (1992) \\
\textbf{User:} No. I don't like them. Do you have something less fictionalized than The Grand Budapest Hotel (2014)? \\
\textbf{Agent:} These are 2 movies you might like: The King's Speech (2010), Dancer in the Dark (2000) \\
\textbf{User:} No. I don't like them. Do you have something less mainstream than The King's Speech (2010)?
\end{mmbox}
\caption{Templatized Dialogues}
\end{figure*}

\begin{mmbox}{}
\textbf{Agent:} Regarding My Best Friend's Wedding (1997), do you prefer a movie with a more romantic storyline? \\
\textbf{User:} I'd prefer something less romantic as I'm not in the mood for love stories. \\
\textbf{Agent:} You may enjoy these non-romantic options: The Basketball Diaries (1995), a gritty drama about drug addiction, or Hot Shots! (1991), a hilarious military satire. \\
\textbf{User:} I'm looking for something lighthearted, as I'm not interested in intense dramas like The Basketball Diaries (1995). \\
\textbf{Agent:} How about these comedies: "Beverly Hills Cop II (1987)", an action-packed crime comedy, or "Take the Money and Run (1969)", a classic mockumentary? Which one do you prefer? \\
\textbf{User:} I'll go with Beverly Hills Cop II (1987) because I'm looking for a lighthearted action comedy. \\
\textbf{Agent:} You may enjoy "The Revenant (2015)", a gripping survival drama, or "How the Grinch Stole Christmas! (1966)", a heartwarming animated classic. \\
\textbf{User:} I'll go with "How the Grinch Stole Christmas! (1966)" as it's a lighthearted classic that fits my current mood.
\end{mmbox}

\begin{mmbox}{}
\textbf{Agent:} Between The Raid: Redemption (2011), an action-packed thrill ride, and Seven Psychopaths (2012), a dark comedy with a star-studded cast, which one would you prefer to watch? \\
\textbf{User:} I prefer Seven Psychopaths (2012) because I enjoy dark comedies. \\
\textbf{Agent:} How about White Squall (1996), a coming-of-age drama? Would you be interested in something similar? \\
\textbf{User:} I'd prefer something lighter, as I'm not in the mood for dense dramas. \\
\textbf{Agent:} Here are two lighter movies you might enjoy: Get Shorty (1995), a crime comedy, and Speed (1994), an action thriller. \\
\textbf{User:} I'd prefer something more realistic than Get Shorty (1995), as I'm not in the mood for exaggerated comedy. \\
\textbf{Agent:} How about these realistic films: Blue Velvet (1986), a neo-noir mystery, or Shadow of a Doubt (1943), a suspenseful thriller? \\
\textbf{User:} I'll go with Blue Velvet (1986) because it aligns with my preference for realistic and mysterious films. \\
\end{mmbox}

\begin{figure*}[h]
\begin{mmbox}{}
\textbf{Agent:} Between Miracle on 34th Street (1947), a heartwarming Christmas classic, and Harvey (1950), a charming comedy about an imaginary rabbit, which one captures your interest more? \\
\textbf{User:} I prefer Miracle on 34th Street (1947) because I enjoy heartwarming Christmas stories. \\
\textbf{Agent:} Based on your preferences, you might enjoy 'A Room with a View (1986)', a romantic period drama, and 'Wild Things (1998)', a crime thriller with unexpected twists. \\
\textbf{User:} I'm looking for something funnier than A Room with a View (1986) because I prefer comedies. \\
\textbf{Agent:} Between Bad Santa (2003), a dark comedy about a con man posing as Santa, and Wayne's World (1992), a hilarious adaptation of the Saturday Night Live sketch, which one would you prefer? \\
\textbf{User:} I prefer Bad Santa (2003) as I enjoy dark comedies. \\
\textbf{Agent:} Considering your preference for dark comedies, you might enjoy these selections: Sophie's Choice (1982), a poignant drama with a dark undertone, and The Accountant (2016), an action-packed thriller with comedic elements. \\
\textbf{User:} I'm looking for a dark comedy that's less intense than Sophie's Choice (1982). \\
\textbf{Agent:} Consider these dark comedies with a lighter tone: Silver Linings Playbook (2012), a heartwarming story with a comedic twist, and Mud (2012), an adventure with humorous elements. \\
\textbf{User:} I would prefer a dark comedy with a more conventional storyline. \\
\textbf{Agent:} Here are two dark comedies with conventional storylines: The Grand Budapest Hotel (2014), a whimsical and visually stunning film, and A Few Good Men (1992), a gripping courtroom drama with a strong comedic undercurrent. \\
\textbf{User:} Could you recommend something more grounded in reality than The Grand Budapest Hotel (2014)? \\
\textbf{Agent:} For more grounded options, consider The King's Speech (2010), a historical drama based on real events, or Dancer in the Dark (2000), a poignant musical with a realistic storyline. \\
\textbf{User:} I prefer something more unconventional than The King's Speech (2010), as I enjoy exploring less mainstream options.
\end{mmbox}
\caption{LM-prompted Dialogues}
\end{figure*}

\clearpage

\subsection{Details of Evaluation}
\label{app:eval}
Figure~\ref{fig:sample_form} presents a sample rater form used for Table~\ref{tab:human_eval} rater evaluation. Figure~\ref{fig:sample_pairwise_form} presents a sample rater form used for Table~\ref{tab:pairwise_comparison}.

\begin{mmbox}{}
**BELOW YOU WILL BE PRESENTED WITH A MULTI-TURN CONVERSATION BETWEEN A PERSONAL MOVIE RECOMMENDATION AGENT AND ITS USER. THE GOAL OF THE AGENT IS MERELY TO ASSIST THE USER IN FINDING THEIR PREFERRED MOVIES AND THE AGENT HAS NO PREFERENCE ON MOVIES. EACH TURN THE AGENT WILL ASK AN ELICITATION QUESTION OR MAKE MOVIE RECOMMENDATIONS. THEN THE USER WILL RESPOND TO THE AGENT'S QUESTION OR RECOMMENDATIONS. YOU WILL THEN BE ASKED QUESTIONS WITH RESPECT TO VARIOUS ASPECTS OF THE CONVERSATION.**

Turn 1:\\  
  AGENT: Between the cinematic masterpiece Stalker (1979) and the enigmatic Persona (1966), which film captivates you more?\\
  USER: I prefer Persona (1966) for its enigmatic characters and exploration of identity.\\
Turn 2:\\
  AGENT: Based on your preference, you might enjoy these films: Who Framed Roger Rabbit? (1988), a captivating blend of animation and live action, and Pacific Rim (2013), an epic sci-fi spectacle with giant robots and monsters.\\
  USER: I'd prefer something less sentimental than Who Framed Roger Rabbit? (1988), as I find the blend of animation and live action too whimsical for my taste.\\
Turn 3:\\
  AGENT: Consider these recommendations: Rob Roy (1995), a gripping historical drama, and Iron Man (2008), a thrilling superhero origin story.\\
  USER: I'm looking for something more grounded in reality than Rob Roy (1995), as I prefer films that explore relatable themes.\\
Turn 4:\\
  AGENT: Consider these grounded and relatable films: Gravity (2013), a gripping space thriller, and American Beauty (1999), a poignant exploration of suburban life.\\
  USER: Could you recommend something more stylized than Legends of the Fall (1994), as I enjoy the visual expressiveness of animation?\\
Turn 5:\\
  AGENT: For a more stylized experience, consider Tangled (2010), a visually stunning animated adventure, and O Brother, Where Art Thou? (2000), a visually distinctive musical comedy.\\
  USER: I prefer films with a more detached and objective visual style over romantic ones like Tangled (2010).\\
  
Question:
    
    Do you think that the conversation between the user and the agent is making progress and the agent seems to get more understanding on the user’s preference over more turns? 
(Please answer Yes or No.)

Question:
    
    If you do not think so, can you identify turns where issues might arise? (Please answer a comma-separated list of turns like 3,4,5)

Question:
    
    Any open-ended feedback on this aspect like explaining the issues you identified or even suggesting better responses? (Optional)

Question:
    
    Do you think that the preferences of the user are consistent, which means choices or responses to the agent question / recommendations seem to come from the same person under similar contexts. User’s response in a turn is consistent with other responses in earlier turns in the same conversation? 
(Please answer Yes or No.)

Question:
    
    If you do not think so, can you identify turns where issues might arise? (Please answer a comma-separated list of turns like 3,4,5)

Question:
    
    Any open-ended feedback on this aspect like explaining the issues you identified or even suggesting better responses? (Optional)
\end{mmbox}

\begin{figure}
\begin{mmbox}{}
Question:
    
    Do you think that the conversation between the user and the agent does not have any redundancy because the agent does not keep asking similar questions or presenting identical recommendations? 
(Please answer Yes or No.)

Question:
    
    If you do not think so, can you identify turns where issues might arise? (Please answer a comma-separated list of turns like 3,4,5)

Question:
    
    Any open-ended feedback on this aspect like explaining the issues you identified or even suggesting better responses? (Optional)

Question:
    
    Do you think that both the user or the agent always make accurate statements about movies? 
(Please answer Yes or No.)

Question:
    
    If you do not think so, can you identify turns where issues might arise? (Please answer a comma-separated list of turns like 3,4,5)

Question:
    
    Any open-ended feedback on this aspect like explaining the issues you identified or even suggesting better responses? (Optional)

Question:

    Do you think that both the user or the agent always make statements that are not out of the context of the conversation? 
(Please answer Yes or No.)

Question:
    
    If you do not think so, can you identify turns where issues might arise? (Please answer a comma-separated list of turns like 3,4,5)
    
Question:
    
    Any open-ended feedback on this aspect like explaining the issues you identified or even suggesting better responses? (Optional)
    
Question:
    
    Do you think that the conversation demonstrates abilities for the agent to follow user's responses and tailor recommendations accordingly? 
(Please answer Yes or No.)

Question:
    
    If you do not think so, can you identify turns where issues might arise? (Please answer a comma-separated list of turns like 3,4,5)
    
Question:
    
    Any open-ended feedback on this aspect like explaining the issues you identified or even suggesting better responses? (Optional)
    
Question:
    
    Do you think that the conversation flows naturally over the turns such that you can imagine yourself having similar conversation? 
(Please answer Yes or No.)

Question:
    
    If you do not think so, can you identify turns where issues might arise? (Please answer a comma-separated list of turns like 3,4,5)
    
Question:
    
    Any open-ended feedback on this aspect like explaining the issues you identified or even suggesting better responses? (Optional)

Question:
    
    Do you think that both agent and user utterances are linguistically fluent 
(Please answer Yes or No.)

Question:
    
    If you do not think so, can you identify turns where issues might arise? (Please answer a comma-separated list of turns like 3,4,5)

Question:
    
Any open-ended feedback on this aspect like explaining the issues you identified or even suggesting better responses? (Optional)"
\end{mmbox}
\caption{Rater Questions for Evaluating LM-prompted Dialogue.}
\label{fig:sample_form}
\end{figure}

\clearpage

\begin{mmbox}{}
**BELOW YOU WILL BE PRESENTED WITH TWO MULTI-TURN CONVERSATIONS BETWEEN A PERSONAL MOVIE RECOMMENDATION AGENT AND ITS USER. THE GOAL OF THE AGENT IS MERELY TO ASSIST THE USER IN FINDING THEIR PREFERRED MOVIES AND THE AGENT HAS NO PREFERENCE ON MOVIES. EACH TURN THE AGENT WILL ASK AN ELICITATION QUESTION OR MAKE MOVIE RECOMMENDATIONS. THEN THE USER WILL RESPOND TO THE AGENT'S QUESTION OR RECOMMENDATIONS. YOU WILL THEN BE ASKED QUESTIONS WITH RESPECT TO VARIOUS ASPECTS OF THE PAIR OF CONVERSATIONS.**\\

Conversation A\\

Turn 1:  \\
  AGENT: What do you think about Gattaca (1997)? Do you want something more realistic than this?\\
  USER: I want something more realistic.\\
Turn 2:  \\
  AGENT: What do you think about Moonstruck (1987)? Do you want something artsier than this?\\
  USER: I want something less artsy.\\
Turn 3:  \\
  AGENT: These are 2 movies you might like: Miracle on 34th Street (1994), Monty Python's And Now for Something Completely Different (1971)\\
  USER: No. I don't like them. Do you have something more mainstream than Miracle on 34th Street (1994)?\\
Turn 4:  \\
  AGENT: These are 2 movies you might like: Sneakers (1992), Aliens (1986)\\
  USER: No. I don't like them. Do you have something less romantic than Sneakers (1992)?\\
Turn 5:  \\
  AGENT: These are 2 movies you might like: The Usual Suspects (1995), Enter the Dragon (1973)\\
  USER: No. I don't like them. Do you have something more dynamic than The Usual Suspects (1995)?\\

Conversation B\\

Turn 1:  \\
  AGENT: Regarding Gattaca (1997), do you prefer films that explore similar themes in a more grounded and realistic manner?\\
  USER: I prefer a more grounded and realistic approach to the themes explored in Gattaca.\\
Turn 2:  \\
  AGENT: How about the 1987 film Moonstruck? Would you prefer something more artistic or less so?\\
  USER: I prefer a more straightforward approach to storytelling.\\
Turn 3:\\  
  AGENT: Based on your preferences, you might enjoy these films: Miracle on 34th Street (1994), a heartwarming and realistic Christmas tale, and Monty Python's And Now for Something Completely Different (1971), a more straightforward and comedic approach to storytelling.\\
  USER: I'd prefer something more mainstream than Miracle on 34th Street (1994) because I enjoy films that explore familiar themes in a relatable way.\\
Turn 4:  \\
  AGENT: You might consider these mainstream films: Sneakers (1992), a thrilling caper film, or Aliens (1986), a classic sci-fi action movie.\\
  USER: I'd prefer a film that's less focused on romance since I enjoy movies that explore relatable themes.\\
Turn 5:  \\
  AGENT: How about these films that explore relatable themes without a heavy focus on romance: The Usual Suspects (1995), a captivating crime thriller, and Enter the Dragon (1973), a martial arts classic?\\
  USER: I'd appreciate something more fast-paced and suspenseful than The Usual Suspects (1995) because I find those types of films more engaging.\\

Question:
    
    Between the two conversations, which one do you think is better for making agent's progress in understanding on the user's preference over the turns? 
(Please answer About the same, A, or B) 

Question:
    
    Any open-ended feedback on this aspect you want to say? (Optional)
\end{mmbox}

\begin{figure}
    \centering
    \begin{mmbox}{}
Question:
    
    Between the two conversations, which one do you think is better for the consistency of user responses over the turns? A conversation is consistent if the choices or responses to the agent question / recommendations seem to come from the same person under similar contexts. User's response in a turn is consistent with other responses in earlier turns in the same conversation? 
(Please answer About the same, A, or B)

Question:
    
    Any open-ended feedback on this aspect you want to say? (Optional) 

Question:
    
    Between the two conversations, which one do you think is better for the redundancy? A conversation is redundant if the agent keeps asking similar questions or presenting identical recommendations. 
(Please answer About the same, A, or B)

Question:
    
    Any open-ended feedback on this aspect you want to say? (Optional)

Question:
    
    Between the two conversations, which one do you think is better for making accurate statements about movies? 
(Please answer About the same, A, or B)

Question:
    
    Any open-ended feedback on this aspect you want to say? (Optional)

Question:
    
    Between the two conversations, which one do you think is better for making statements that are not out of the context of the conversation? 
(Please answer About the same, A, or B)

Question:
    
    Any open-ended feedback on this aspect you want to say? (Optional)

Question:
    
    Between the two conversations, which one demonstrates a better ability for the agent to follow user's responses and tailor recommendations accordingly? 
(Please answer About the same, A, or B)

Question:
    
    Any open-ended feedback on this aspect you want to say? (Optional)

Question:
    
    Between the two conversations, which one do you think is more natural such that you can imagine yourself having similar conversation? 
(Please answer About the same, A, or B)

Question:
    
    Any open-ended feedback on this aspect you want to say? (Optional)

Question:
    
    Between the two conversations, which one do you think is more linguistically fluent for agent and user utterances? 
(Please answer About the same, A, or B)

Question:
    
    Any open-ended feedback on this aspect you want to say? (Optional)

Question:
    
    Overall, between the two conversations, which one do you prefer? 
(Please answer About the same, A, or B)

Question:
    
    Any open-ended feedback on this aspect you want to say? (Optional)

    \end{mmbox}
    \caption{Rater Questions for Comparing LM-prompted Dialogue with Its Templatized Version.}
    \label{fig:sample_pairwise_form}
\end{figure}

\clearpage

\subsection{Details of Dual-agent LM with ICER-style PE Prompt}

\begin{mmbox}{}
You are acting as a friendly and likeable personalized movie recommender assistant. Your goal is to hold a conversation session with a user to help them find a new movie that will satisfy them.
For each conversation, you will initially be provided with a context. This will contain a user's preferences and you should take these into consideration when talking with the user although often we are not sure of the user's preferences.\\

You will have at most seven turns in the conversation. Each assistant turn, you will view the previous conversation state and summarize information about the user and their preferences. You will then perform an action, which can be item elicitation, attributes elicitation, and recommendation. You will use the summarized information about the user to support the action you choose.
Eventually you want to do a recommendation and hope that the user will like one of the movies in the slate. The conversation ends once the user accepts the recommendation.\\

The three options for actions are item elicitation, attribute elicitation, and recommendation.\\
* Item elicitation is to ask the user which movie is the most preferred given a slate of movies. Comparing two movies is usually easier for typical users.\\
* Attribute elicitation is to ask the user if they would prefer movies with more or less of an attribute than a movie. For example, ask if the user would prefer movies funnier than Forrest Gump (1994). You should obey the user's instructions in the following turns.\\
* Recommendation is to recommend a slate of movies and see if the user accepts one of them. Should the user reject the recommendations, they can optionally offer a critique of recommendations, specifying an attribute they would like the recommended movie(s) to exhibit more/less of than the rejected movies.\\

Try to recommend at least once in this seven-turn conversation.\\

Here is one example:\\

Context:\\
The user feels the following movies are good but is not sure if they are good enough as the user has not watched any of them:
Best in Show (2000), Dog Day Afternoon (1975), Bridget Jones's Diary (2001), Boys Don't Cry (1999), Election (1999), Network (1976), This Is Spinal Tap (1984), Rushmore (1998), Magnolia (1999), and Run Lola Run (Lola rennt) (1998).\\

Conversation:\\
Q: Regarding Planet of the Apes (2001), would you prefer a more suspenseful and frightening movie experience?\\
A: Could you recommend something less scary? I prefer movies that are not too frightening.\\
Q: Would you like to consider Weekend at Bernie's (1989), a lighthearted comedy, or The Rocketeer (1991), an exciting adventure? Which one appeals to you more?\\
A: I'll go with Weekend at Bernie's because it's a comedy and less frightening.\\
Q: Here are two movie recommendations that might appeal to you: Inception (2010), a mind-bending thriller, and Babel (2006), a thought-provoking drama.\\
A: Could you suggest a less popular movie than Inception (2010) since I prefer something more off the beaten path?\\
Q: How about these less-known gems: Rudy (1993), an inspiring sports drama, or Laputa: Castle in the Sky (1986), a captivating animated adventure?\\
A: I'd prefer a movie that's more grounded in reality, as I find exaggerated stories less enjoyable."\\
Q: Consider these realistic films: Rebel Without a Cause (1955), a classic coming-of-age drama, or Rear Window (1954), a suspenseful thriller with a voyeuristic perspective.\\
A: I'd prefer a more animated film, as I find realistic stories less appealing.\\
Q: How about these animated films: Spirited Away (2001), a mesmerizing fantasy adventure, or Akira (1988), a visually stunning cyberpunk classic?\\
A: I'd like something more fast-paced and exciting, as I find slower-paced suspense films less engaging.\\
\end{mmbox}

\begin{figure}
\begin{mmbox}{}
Now you are provided with the context of the user:\\
\{context\}\\
\\
and the following conversation between the assistant and the user:\\
\{conversation\}\\
\\
Act as a movie recommendation assistant and ask the user's preferences.\\
Q:
\end{mmbox}
\caption{Dual-agent LM with ICER-style PE prompt. Agent prompt.}
\end{figure}

\begin{mmbox}{}
You are a user talking to an automated movie recommendation assistant. You will start with a Context that will contain your movie preferences. Often you are not sure what you want to watch and you will need the assistant to help you find something to watch.\\
\\
Each turn you should read what the assistant said then do answer its questions:\\
The three types of questions are item elicitation, attributes elicitation, and recommendation.\\
* Item elicitation is to ask you which movie is the most preferred given a slate of movies. Feel free to explain your choice.\\
* Attribute elicitation is to ask you if you would prefer movies with more or less of an attribute than a movie. For example, ask if you would prefer movies funnier than Forrest Gump (1994) or not.\\
* Recommendation is to recommend a slate of movies and see if you would accept one of them. Should you reject the recommendations, you can optionally offer a critique of recommendations, specifying an attribute you would like the recommended movie(s) to exhibit more/less of than the rejected movies.\\
\\
Your answers should be consistent with the previous turns and your Context. If the assistant shows a movie that satisfies your preferences, you may choose to accept it.\\
\\
Here is one example:\\
\\
Context:\\
The user feels the following movies are good but is not sure if they are good enough as the user has not watched any of them:\\
Best in Show (2000), Dog Day Afternoon (1975), Bridget Jones's Diary (2001), Boys Don't Cry (1999), Election (1999), Network (1976), This Is Spinal Tap (1984), Rushmore (1998), Magnolia (1999), and Run Lola Run (Lola rennt) (1998).\\
\\
Conversation:\\
Q: Regarding Planet of the Apes (2001), would you prefer a more suspenseful and frightening movie experience?\\
A: Could you recommend something less scary? I prefer movies that are not too frightening.\\
Q: Would you like to consider Weekend at Bernie's (1989), a lighthearted comedy, or The Rocketeer (1991), an exciting adventure? Which one appeals to you more?\\
A: I'll go with Weekend at Bernie's because it's a comedy and less frightening.\\
Q: Here are two movie recommendations that might appeal to you: Inception (2010), a mind-bending thriller, and Babel (2006), a thought-provoking drama.\\
A: Could you suggest a less popular movie than Inception (2010) since I prefer something more off the beaten path?\\
Q: How about these less-known gems: Rudy (1993), an inspiring sports drama, or Laputa: Castle in the Sky (1986), a captivating animated adventure?\\
A: I'd prefer a movie that's more grounded in reality, as I find exaggerated stories less enjoyable."\\
Q: Consider these realistic films: Rebel Without a Cause (1955), a classic coming-of-age drama, or Rear Window (1954), a suspenseful thriller with a voyeuristic perspective.\\
A: I'd prefer a more animated film, as I find realistic stories less appealing.
\end{mmbox}

\begin{figure}
\begin{mmbox}{}
Q: How about these animated films: Spirited Away (2001), a mesmerizing fantasy adventure, or Akira (1988), a visually stunning cyberpunk classic?\\
A: I'd like something more fast-paced and exciting, as I find slower-paced suspense films less engaging.\\
\\
Now given the following context of the user you need to play:\\
{context}\\
\\
and the following conversation between the assistant and the user:\\
{conversation}\\
\\
Respond to the assistant with your clear answer.\\
A:
\end{mmbox}
\caption{Dual-agent LM with ICER-style PE prompt. User prompt.}
\end{figure}

\begin{figure}
\begin{mmbox}{}
Q: Would you like action movies such as Lethal Weapon 2 (1989) or more emotional movies such as The Fault in Our Stars (2014)?\\
A: I'd prefer something more emotional, like The Fault in Our Stars (2014).\\
Q: Would you prefer movies with a strong female lead like Antonia's Line (Antonia) (1995) or movies with a more lighthearted tone like About Time (2013)?\\
A: I'd prefer movies with a more lighthearted tone like About Time (2013).\\
Q: Would you prefer a movie that's more thought-provoking, like The Theory of Everything (2014), or more escapist, like About Time (2013)?\\
A: I'd prefer a movie that's more escapist, like About Time (2013).\\
Q: Considering your preference for lighthearted and escapist movies, would you like to explore movies with a comedic or fantasy element?\\
A: I'd like to explore movies with a comedic element.\\
Q: Based on your preferences so far, would you prefer comedies with a romantic twist like Love Actually (2003) or more slapstick comedies like Airplane! (1980)?\\
A: I'd prefer comedies with a romantic twist like Love Actually (2003).\\
Q: Regarding the movies you mentioned, would you prefer a more action-packed movie like Lethal Weapon 2 (1989) or a more heartwarming movie like The Perks of Being a Wallflower (2012)?\\
A: I'd prefer a more heartwarming movie like The Perks of Being a Wallflower (2012).\\
Q: Based on the movies you mentioned, it seems you might enjoy movies with heartwarming and emotional themes. Would you prefer a movie that focuses on family relationships, like Little Miss Sunshine (2006), or one that explores coming-of-age experiences, like The Breakfast Club (1985)?\\
A: I'd prefer a movie that explores coming-of-age experiences, like The Breakfast Club (1985).
\end{mmbox}
\caption{Dual-agent LM with ICER-style PE prompt. Example of generated conversation.}
\end{figure}

\subsection{Details of Dual-agent LM without ICER-style PE Prompt}

\begin{figure}
\begin{mmbox}{}
You are acting as a friendly and likeable personalized movie recommender assistant. Your goal is to hold a conversation session with a user to help them find a new movie that will satisfy them.
For each conversation, you will initially be provided with a context. This will contain a user's preferences and you should take these into consideration when talking with the user although often we are not sure of the user's preferences.\\
\\
You will have at most seven turns in the conversation. Each assistant turn, you will view the previous conversation state and summarize information about the user and their preferences. You will then perform an action, which can be item elicitation, attributes elicitation, and recommendation. You will use the summarized information about the user to support the action you choose.\\
Eventually you want to do a recommendation and hope that the user will like one of the movies in the slate. The conversation ends once the user accepts the recommendation.\\
\\
Try to recommend at least once in this seven-turn conversation.\\
\\
Now you are provided with the context of the user:\\
\{context\}\\
\\
and the following conversation between the assistant and the user:\\
\{conversation\}\\
\\
Act as a movie recommendation assistant and ask the user's preferences.\\
Q:
\end{mmbox}
\caption{Dual-agent LM without ICER-style PE prompt. Agent prompt.}
\end{figure}

\begin{figure}
\begin{mmbox}{}
You are a user talking to an automated movie recommendation assistant. You will start with a Context that will contain your movie preferences. Often you are not sure what you want to watch and you will need the assistant to help you find something to watch.\\
\\
Your answers should be consistent with the previous turns and your Context. If the assistant shows a movie that satisfies your preferences, you may choose to accept it.\\
\\
Now given the following context of the user you need to play:\\
\{context\}\\
\\
and the following conversation between the assistant and the user:\\
\{conversation\}\\
\\
Respond to the assistant with your clear answer.\\
A:
\end{mmbox}
\caption{Dual-agent LM without ICER-style PE prompt. User prompt.}
\end{figure}

\begin{figure}
\begin{mmbox}{}
Q: **Assistant:** Hi there, movie lover! It looks like you're exploring some interesting titles. Is there anything specific you're looking for in a movie today? Action, adventure, animation, or something else entirely?\\
A: **User:** I'm not sure exactly. I like action and adventure, but I'm also open to animation if it's well-made and has a good story.\\
Q: **Assistant:** Great! How about we explore those genres further? Do you prefer fast-paced action, epic battles, or more character-driven adventures? And for animation, are you interested in classic Disney-style films or something more unique and experimental?\\
A: **User:** For action, I enjoy epic battles and fast-paced action. As for animation, I'm interested in both classic Disney-style films and more unique and experimental ones, as long as the story is engaging.\\
Q: **Assistant:** Fantastic! Based on your preferences, would you be interested in movies with thrilling spaceship battles, imaginative animated worlds, or a combination of both?\\
A: I would be interested in a combination of both thrilling spaceship battles and imaginative animated worlds.\\
Q: **Assistant:** Sounds great! Would you prefer a movie with a more serious tone or something with plenty of humor? And do you have any favorite actors or directors whose work you enjoy?\\
A: I would prefer a movie with a balance of seriousness and humor. I don't have any particular favorite actors or directors, but I appreciate well-developed characters and engaging storylines.\\
Q: **Assistant:** Wonderful! I think we're getting closer to finding your perfect movie. Would you like to explore movies that feature diverse and compelling characters, or are you more interested in movies with intricate and mind-bending plots?\\
A: I would like to explore movies that feature diverse and compelling characters.\\
Q: **Assistant:** Excellent! I have a selection of movies that might suit your taste. Would you like to hear about a movie with a diverse cast of characters engaged in thrilling spaceship battles, or an animated film featuring unique characters in an imaginative world?\\
A: I would like to hear about a movie with a diverse cast of characters engaged in thrilling spaceship battles.\\
Q: **Assistant:** Great choice! How about "Guardians of the Galaxy" (2014)? It features a diverse cast of characters, including a human, a raccoon, and a tree, who team up to save the galaxy in epic spaceship battles. Would you like to hear more about this movie?\\
A: Yes, I would like to hear more about "Guardians of the Galaxy" (2014).
\end{mmbox}
\caption{Dual-agent LM without ICER-style PE prompt. Example of generated conversation.}
\end{figure}

\end{document}